\pdfoutput=1

\documentclass[11pt]{article}

\usepackage{acl}
\usepackage{multirow}
\usepackage{times}
\usepackage{latexsym}
\usepackage{booktabs}
\usepackage{amsmath}
\usepackage{graphicx}
\usepackage{float}
\usepackage{wrapfig}
\usepackage{xcolor}
\usepackage{array}
    \newcolumntype{P}[1]{>{\centering\arraybackslash}p{#1}}
    \newcolumntype{M}[1]{>{\centering\arraybackslash}m{#1}}

\usepackage[T1]{fontenc}

\usepackage[utf8]{inputenc}

\usepackage{microtype}

\newif\ifhidecomments

\ifhidecomments
\usepackage{environ}
\NewEnviron{hide}{}

\fi

\title{Time is Encoded in the Weights of Finetuned Language Models}

\author{
    \bf Kai Nylund$^{1}$ \quad
	\bf Suchin Gururangan$^{1}$ \quad
	\bf Noah A. Smith$^{1, 2}$ \quad \\
	$^1$Paul G. Allen School of Computer Science \& Engineering, University of Washington \\
	$^2$Allen Institute for AI \\
	{\tt knylund@cs.washington.edu }
}
\begin{document}
\maketitle

\begin{abstract}

We present \emph{time vectors}, a simple tool to customize language models to new time periods. Time vectors are created by finetuning a language model on data from a single time (e.g., a year or month), and then subtracting the weights of the original pretrained model. This vector specifies a direction in weight space that, as our experiments show, improves performance on text from that time period. Time vectors specialized to adjacent time periods appear to be positioned closer together in a manifold. Using this structure, we interpolate between time vectors to induce new models that perform better on intervening and future time periods, without any additional training. We demonstrate the consistency of our findings across different tasks, domains, model sizes, and time scales. Our results suggest that time is encoded in the weight space of finetuned models. 
\end{abstract}

\section{Introduction}

Temporal variation is a fundamental characteristic of language. As we show in \S\ref{sec:temporal_misalignment}, it manifests in language model development as \emph{temporal misalignment}, where deviations in train and test data lead to large performance degradation across different time periods \citep[\emph{inter alia}]{luu-etal-2022-time, lazaridou2021mind, jaidka2018diachronic}. 
This necessitates adaptation techniques for customizing models to specific time periods as needed.
Designing such techniques is difficult, however, due to the multitude of time scales and the possibility that data from a target time period might be unavailable.

Recent work has shown that the behavior of neural networks can be edited through closed-form interpolation between parameters of finetuned models \citep[\emph{inter alia}]{ilharco2023editing, OrtizJimnez2023TaskAI, li2022branchtrainmerge, Wortsman2021RobustFO}. In this work, we demonstrate that weight-space interpolation can also be used to cheaply edit language model behavior over \emph{time}. To this end, we introduce \emph{time vectors} (\S\ref{sec:time_vecs}), an extension of task vectors \citep{ilharco2023editing}. We finetune a pretrained language model on text from a single time period, and then subtract the pretrained weights. This vector represents a direction of movement in weight space that improves performance on text from the target time period. 

\begin{figure}[t!]
    \centering
    \includegraphics[width=\columnwidth]{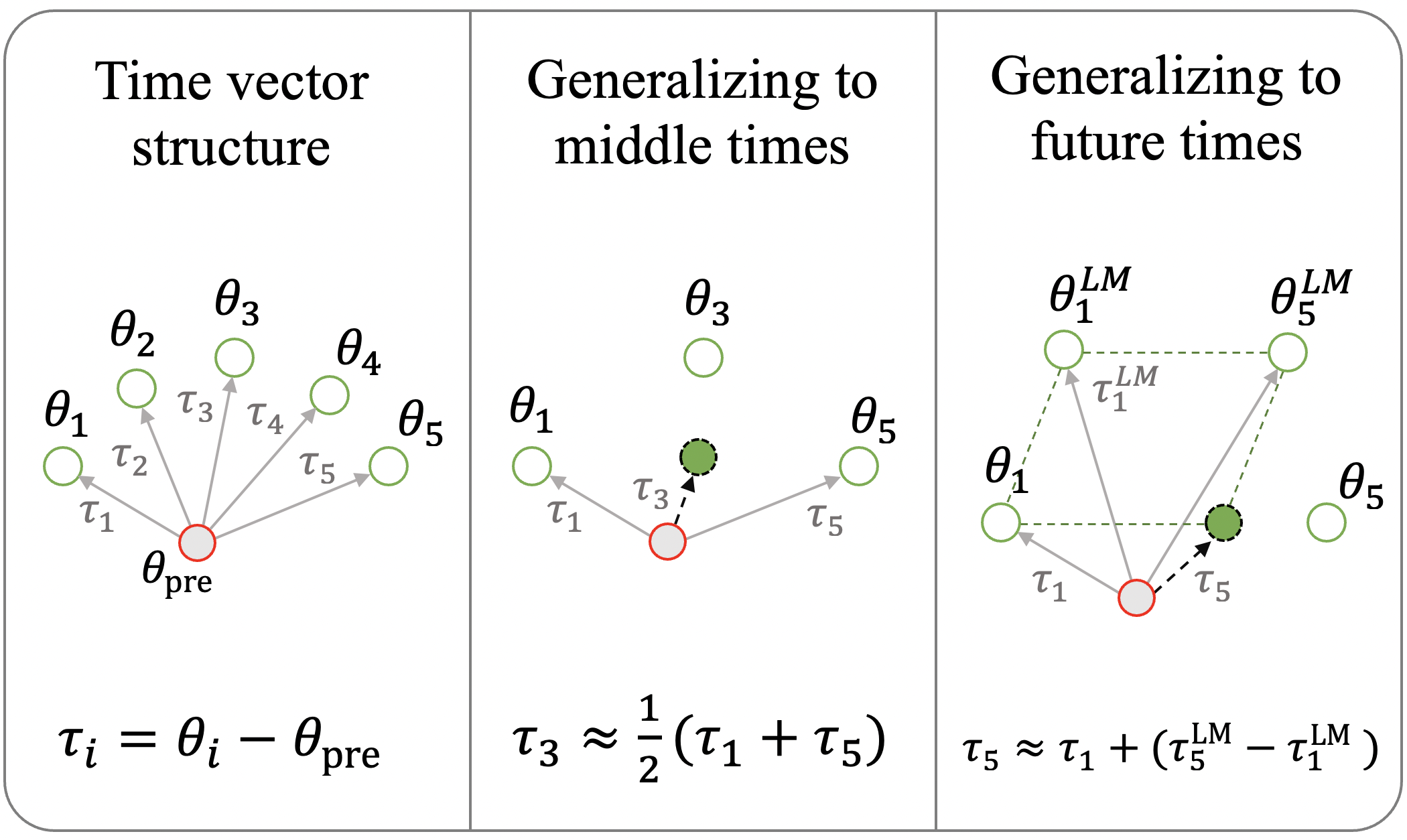}
    \caption{\textbf{We present \emph{time vectors}, a simple tool to customize language models to new time periods.} Time vectors ($\tau_i$)  specify a direction in weight space that improves performance on text from a time period $i$. They are computed by subtracting the pretrained weights ($\theta_{\text{pre}}$; left panel) from those finetuned to a target time period ($\theta_i$). We can customize model behavior to new time periods (e.g., intervening months or years) by interpolating between time vectors and adding the result to the pretrained model (middle panel). We can also generalize to a future time period $j$ with analogy arithmetic (right panel). This involves combining a task-specific time vector with analogous time vectors derived from finetuned language models ($\tau^{\text{LM}}_j$).}
    \label{fig:time_vecs}
\end{figure}

\begin{figure*}[t!]
    \centering
    \includegraphics[width=\textwidth]{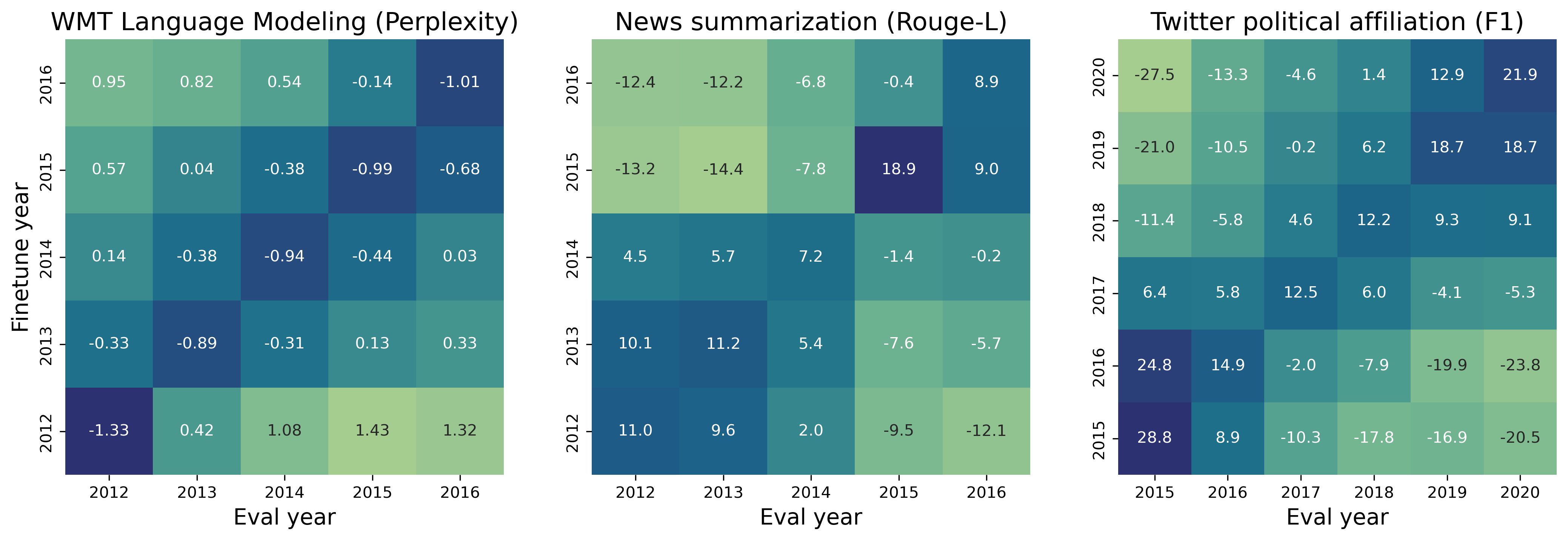}
    \caption{\textbf{Model performance degrades linearly year-to-year.}  We evaluate language model perplexity (WMT), ROUGE-L (news summarization), and macro F1 (political affiliation classification). Each cell indicates the monthly performance of T5-3B finetuned and evaluated on a \emph{single year} from that task. We report the percentage difference from the average performance for each year, and find linear degradation as finetuning and evaluation years become more misaligned regardless of task. We display similar trends for T5-small and medium, as well as for other domains and tasks, in \S\ref{subsec:other_yearly_misalignment}. We measure the linearity of these degradations in Appendix Table \ref{table:td_scores}.}
    \label{fig:yearly_misalignment}
\end{figure*}

We analyze the structure of time vectors with temporally organized datasets for language modeling, classification, and summarization (\S\ref{sec:data}). Our results consistently suggest that time vectors are intuitively organized on a manifold; years or months that are closer together in time yield time vectors that are also closer together in weight space. Similarly, we show that temporal degradation in yearly and monthly settings is strongly correlated with the angles between time vectors (\S\ref{subsec:cos_sim_static}).

We use this structure of time vectors to induce models that generalize better to data from new time periods.  By interpolating between two time vectors, we discover vectors that, when applied to the pretrained model, improve performance on intervening months or years (\S\ref{subsec:intervening}). 
The structure can also be used to generalize task-specific models across time periods with analogous time vectors specialized to unlabeled data (\S\ref{subsec:task_analogy}).  

Our results show that temporal variation is to some extent encoded in the weight space of finetuned models, and that weight interpolation can help customize language models to new time periods. We publicly release our code, data, and over 500 models finetuned on specific time periods.\footnote{\url{https://github.com/KaiNylund/lm-weights-encode-time}}

\section{Data and Finetuning}
\label{sec:data}

In this section, we describe our datasets and finetuning techniques, which serve as the basis for all subsequent experiments. We finetune language models on multiple time-stratified datasets, which we use to analyze temporal misalignment and build time vectors. Then, we explore different ways of interpolating between time vectors to generalize to new times. See \S\ref{subsec:intervening}-\ref{subsec:multi-year} for more details on interpolation strategies.

\subsection{Datasets}

\paragraph{Language Modeling} We create two new time-specific language modeling datasets from unlabeled text in news and Twitter domains. For these datasets, we measure \emph{perplexity} of the model on the test set:

\begin{itemize}
    \item \underline{WMT Language Modeling}:
     We randomly sample 67K $\pm$ 5K articles (47M BPE tokens) of training articles and 3K $\pm$ 0.3K test articles (2.3--2.4M tokens of) from each year 2012--2021 in the English subset of the WMT news dataset \cite{wmt-2021-machine}, from 2012--2016. From the same time range, we also sample 7.1M tokens of training articles and 700--720K tokens of test articles from each month. We are missing WMT train and test splits for August 2012 and May 2016.
        
    \item \underline{Twitter Language Modeling}:
    We randomly sample 2M $\pm$ 105K training tweets (72--78M tokens BPE tokens) and 100K $\pm$ 5.4K test tweets (3.6-3.9M BPE tokens) from each year in the Internet Archive Twitter Stream Grab ,\footnote{\url{https://archive.org/details/twitterstream}} from 2015--2020. We only use this dataset to study the domain-specificity of time vectors in \S\ref{subsec:task_analogy}.

        %
\end{itemize}

To understand the level of contamination in our datasets, we measure the overlap between yearly train and test splits in both tasks using a Bloom filter.\footnote{\url{https://github.com/allenai/bff}} We find that less than two percent and 0.1 percent of examples in the Twitter and WMT LM test sets, respectively, contain contaminated n-grams.

\paragraph{Downstream Tasks}

For downstream tasks, we draw from \citet{luu-etal-2022-time}. We measure each model's performance on the test set in \emph{ROUGE-L} for NewsSum and \emph{macro F1} for PoliAff.

\begin{itemize}
\item \underline{NewsSum}: We use \citet{luu-etal-2022-time}  postprocessing of \citet{grusky2018newsroom} news summarization task. To align with out WMT dataset, we do not bin adjacent years together, creating uniformly sized splits for each year from 2012 to 2016.

   
\item \underline{PoliAff}: We use the Political Affiliation task from \citet{luu-etal-2022-time}, with uniformly sized datasets for each year from 2015 to 2020.

\end{itemize}

\subsection{Finetuning}

To compare the same weight space across tasks, we use pretrained T5 \citep{raffel2023exploring} checkpoints for all our experiments. We finetune T5-small, T5-large, and T5-3b on each of our time-stratified datsets. For language modeling, we use the ``LM adaptation'' objective \citep{lester2021power}.

To reduce the computational burden, we finetune T5-large and T5-3B with Low-Rank Adaptation \citep[LoRA;][]{hu2021lora}  and default hyperparameters ($q$ and $v$ attention target modules, $r = 8, \alpha = 32$, dropout = 0.1). When creating time vectors, we merge LoRA weights back into the base model before subtracting the pretrained model. 

Across all settings, we use a batch size of 2 with 8 gradient accumulation steps. We finetune for a single epoch on LM splits and three epochs on downstream task splits. Our learning rates across all tasks are $8 \times 10^{-4}$ for T5-small and T5-large, and $2 \times 10^{-4}$ for T5-3b. We finetuned models concurrently with a single GPU each; we used 8 2080ti, 4 Titan, and 8 A40 GPUs. In experiments for sections \S\ref{subsec:task_analogy} and \S\ref{subsec:multi-year}, we ran evaluations in parallel using available Titan, A40, and A100 GPUs.

\section{Revealing Temporal Misalignment at Multiple Time Scales}
\label{sec:temporal_misalignment}

We begin with an analysis of temporal misalignment using the new set of models and tasks that we consider in this work (\S\ref{sec:data}). These findings set the stage for our creation of time vectors in \S\ref{sec:time_vecs}.

\subsection{Yearly Degradation is Linear}

Previous work on temporal misalignment shows that models degrade over time on a yearly basis. To confirm these results, we finetune T5-small, T5-large, and T5-3b on each yearly split from every dataset. We then evaluate each of these year-finetuned models on every other time split of the test data. 

We display heatmaps of temporal misalignment at a yearly scale in Figure \ref{fig:yearly_misalignment}. We report percent perplexity change from the average on each year to avoid inherent year performance differences. Consistent with past work \citep{lazaridou2021mind, luu-etal-2022-time, longpre2023pretrainer}, we observe linear patterns of degradation in each task for all model sizes (see Table \ref{table:td_scores} in the Appendix for more details). Like \citet{luu-etal-2022-time} show, some tasks, like political affiliation classification, exhibit clearer degradation than others. We quantify these variations in \S\ref{subsec:linear_deg_variations}.

\begin{figure}[t!]
    \centering
    \includegraphics[width=\columnwidth]{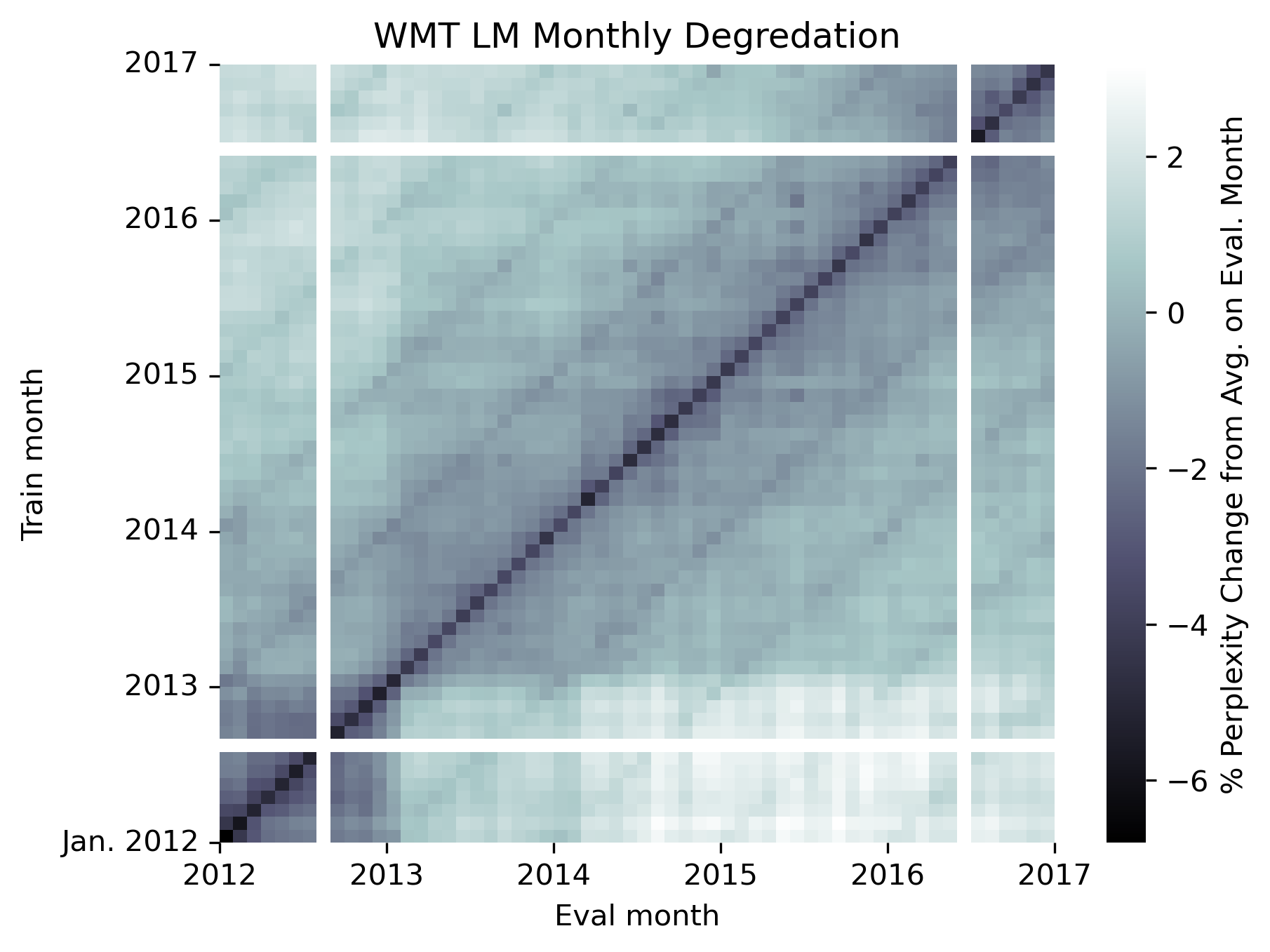}
    \caption{\textbf{Monthly temporal degradation has seasonal patterns.} Each cell indicates the monthly performance of T5-small finetuned and evaluated on a \emph{single month} of the WMT dataset. We report the percentage difference in test perplexity from the average on the evaluation month over all finetuned T5-small models (darker is better). The diagonal indicates that each model does best on its finetuning month. Models also do relatively better on the same month in other years, visible as the stripes radiating out from the diagonal every 12 months.}
    \label{fig:monthly_misalignment}
\end{figure}

\subsection{Monthly Degradation is Seasonal}
\label{subsec:monthly_seasonal}
Next, we turn to month-by-month temporal misalignment, which, to the best of our knowledge, is unexplored. We train T5-small on each WMT LM month split from 2012--2016, resulting in 58 month-finetuned models. We then test every 2012--2016 month model on each month test split for a total of 3,364 evaluations.

As seen in Figure \ref{fig:monthly_misalignment}, finetuning and evaluating models on specific months in the WMT dataset reveals non-linear patterns in temporal misalignment, which correspond to the cycle of months in each year. This pattern is captured by the stripes that occur parallel to the diagonal every 12 months, which indicate that the model for a particular month tends to do better on the same month in other years. We quantify these differences in perplexity in appendix Figure \ref{fig:monthly-violins}. We also report degradation patterns in \emph{online} training settings in \S\ref{subsec:online}. 

\subsection{Summary}

We measure temporal misalignment across a variety of domains, tasks and time scales. While performance decays linearly on a yearly scale, we discover seasonal trends in month-to-month misalignment. Next, we analyze how these phenomena relate to the weights of time-specific models, and then use that relationship to present techniques for adapting LMs to new times.

\section{Temporal Adaptation with Time Vectors}
\label{sec:time_vecs}

\begin{figure}[t!]
    \centering
    \includegraphics[width=\columnwidth]{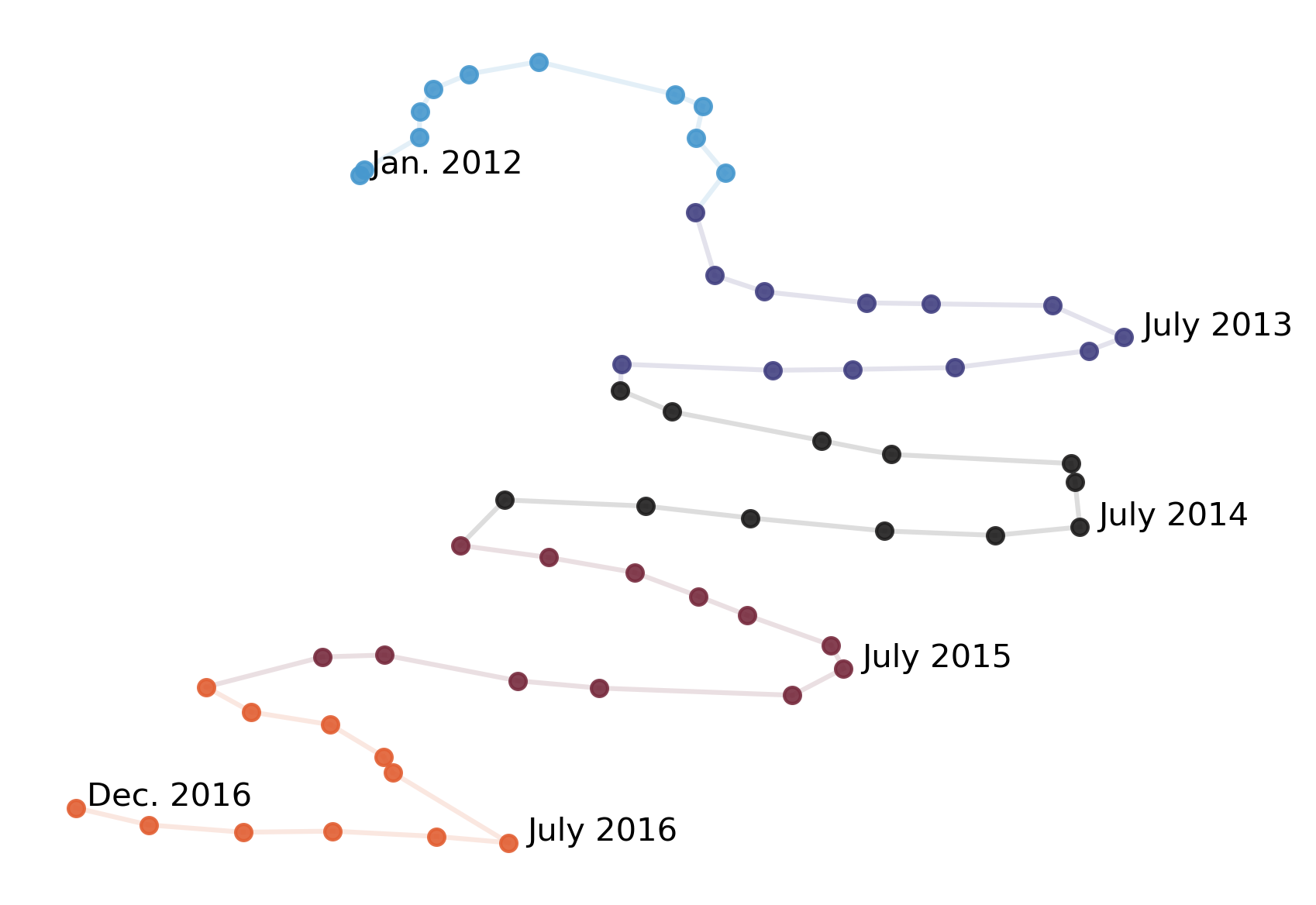}
    \caption{\textbf{Time vectors are organized in a manifold that reflects temporal variation.} Each point is a UMAP projection of the last feedforward layer of a T5-small time vector finetuned on single month of WMT. Points and edges between adjacent months are colored by year. Distances between the weights of time vectors correlate with temporal misalignment (\S\ref{subsec:cos_sim_static}).}
    \label{fig:monthly_time_vecs}
\end{figure}

The collection of year and month-finetuned models from \S\ref{sec:temporal_misalignment} presents a new source of data to study temporal misalignment: model weights. In this section, we analyze these weights through the lens of \emph{time vectors}, formed by taking the difference of a model finetuned on a specific time and the pretrained model. First, we show that the weights of two time vectors become less similar as the times they were finetuned on become more misaligned (\S\ref{subsec:cos_sim_static}). Then, we attempt to use the reverse relationship to update models to unseen times: reducing misalignment on intervening (\S\ref{subsec:intervening}), future (\S\ref{subsec:task_analogy}), and multiple time periods (\S\ref{subsec:multi-year}) by interpolating time vectors. 

\subsection{Background and Definition}

Task vectors \citep{ilharco2023editing} are the difference of the weights of a pretrained model from the weights of the same model after finetuning on a task. Adding and subtracting task vectors from finetuned models is a simple and effective way to improve performance on other settings, or reduce unwanted behavior without further training. Like word embeddings, if there are tasks with the analogous relationship ``$A$ is to $B$ as $C$ is to $D$,'' then task vectors can be used to improve performance on $D$ with the approximation $D \approx C + (B - A)$.

Time vectors are an extension of task vectors to the time domain. Given the weights of the pretrained model, $\theta_{\text{pre}}$ and those of the model finetuned on data from only a single time period $t$, $\theta_t$, a time vector $\tau_t = \theta_t - \theta_{\text{pre}}$ .  Like their task-based counterparts, we add back the pretrained weights at inference time and evaluate $\theta_{\text{pre}} + \tau_t$ \cite{ilharco2023editing}. We call time vectors from models finetuned on individual years and months ``year-vectors'' and ``month-vectors.''

\subsection{Correlation of Time Vector Similarity and Temporal Degradation}
\label{subsec:cos_sim_static}

\begin{table}[t!]
    \small
    \resizebox{\columnwidth}{!}{%
    \begin{tabular}{l c c c c c c c}
        & & \emph{Pearson r} & \\

        \toprule

        T5 size & WMT LM & NewsSum & PoliAff \\
        \midrule
        small & -0.867 & 0.663 & 0.654 \\
        large & -0.737 & 0.628 & 0.672 \\
        3b & -0.795 & 0.626 & 0.668 \\
        \bottomrule
    \end{tabular}}
    \caption{\textbf{The similarity between time vectors correlates  with temporal degradation.} Pearson correlation between cosine similarity of yearly time vectors and \% degradation from the mean performance of all yearly models on each evaluation time period. All $p$-values are $ < 8\times 10^{-4}$.}
    \label{table:perf_cos_sim_correlation}
\end{table}

We visualize time vectors with a UMAP in Figure \ref{fig:monthly_time_vecs}, which suggests that time vectors closer together in weight space are also closer together in time. To verify this hypothesis, we measure the cosine similarity between model weights from each pair of time vectors trained on different time periods (visualized in \S\ref{subsec:other_yearly_misalignment}). 

We find that this similarity metric and performance (Figure \ref{fig:yearly_lm_decay_big}) decay similarly over time. Table \ref{table:perf_cos_sim_correlation} shows that the correlation between cosine similarity and relative performance change on different years is highest in WMT language modeling.  Correlations are generally similar across T5 sizes, with a higher score for T5-small in the WMT LM setting than T5-large and T5-3b, and no absolute values less than 0.6.

This relationship also extends to the monthly scale. Seasonal stripes are  visible in the cosine similarities between each pair of monthly WMT time vectors (visualized in Appendix Figure \ref{fig:monthly_cos_sims}). The monthly performance degradation from the mean (Figure \ref{fig:monthly_misalignment}) and cosine similarity matrices (Figure \ref{fig:monthly_cos_sims}) have a negative correlation (Pearson $r= -0.667$; $p < 10^{-16}$). We analyze cosine similarities to single-year time vectors throughout online training in Appendix \S\ref{subsec:cos_sims_online}.

These results indicate that time vectors are organized in way that is predictive of their performance on corresponding time periods. Next, we explore how we can use this structure to improve performance on new time periods by interpolating between time vectors.

\begin{figure*}[t!]
    \centering
    \includegraphics[width=\textwidth]{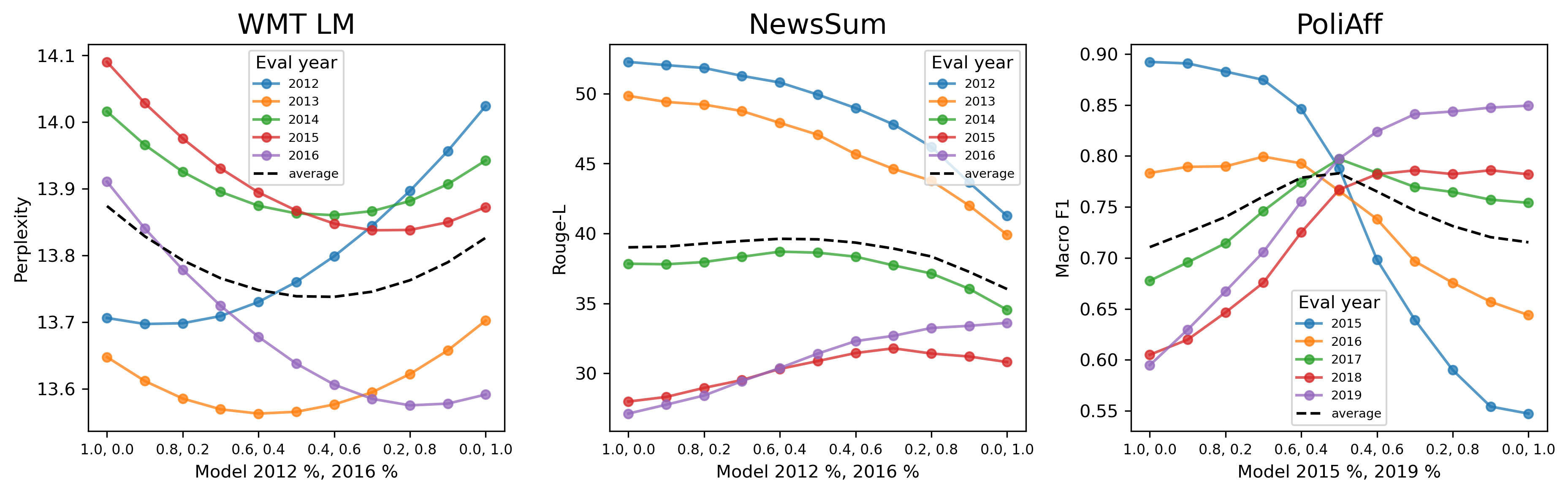}
    \caption{\textbf{Interpolating between two year vectors improves performance on the years between them.} These performance improvements follow an intuitive structure, e.g. when interpolating between 2012 and 2016, the best result on  2013 occurs with a higher percentage of 2012 and vice versa for 2015. Improvement from interpolation varies across settings.}
    \label{fig:intervening_interpolation_years}
\end{figure*}

\subsection{Generalizing to Intervening Time Periods}
\label{subsec:intervening}

Archiving issues or a low sampling rate can lead to gaps in datasets between the oldest and newest examples. Without data, we expect models to perform worse on these ``gap'' times due to temporal misalignment. In this section, we find that we can generalize better to these intervening time periods by interpolating between models finetuned on the oldest and newest times. 

\paragraph{Method} For two time vectors $\tau_{j}$, $\tau_{k}$, we compute their interpolation $\alpha \cdot \tau_{j} + (1 - \alpha)\cdot \tau_{k}$ with $\alpha \in [0, 1]$. In this section, we interpolate between the earliest year time vector $\tau_0$ and latest year time vector $\tau_n$ and evaluate on times $t_0, ... , t_n$ for each $\alpha \in [0.1, 0.2, ..., 1.0]$.

\begin{figure*}[t!]
    \centering
    \includegraphics[width=\textwidth]{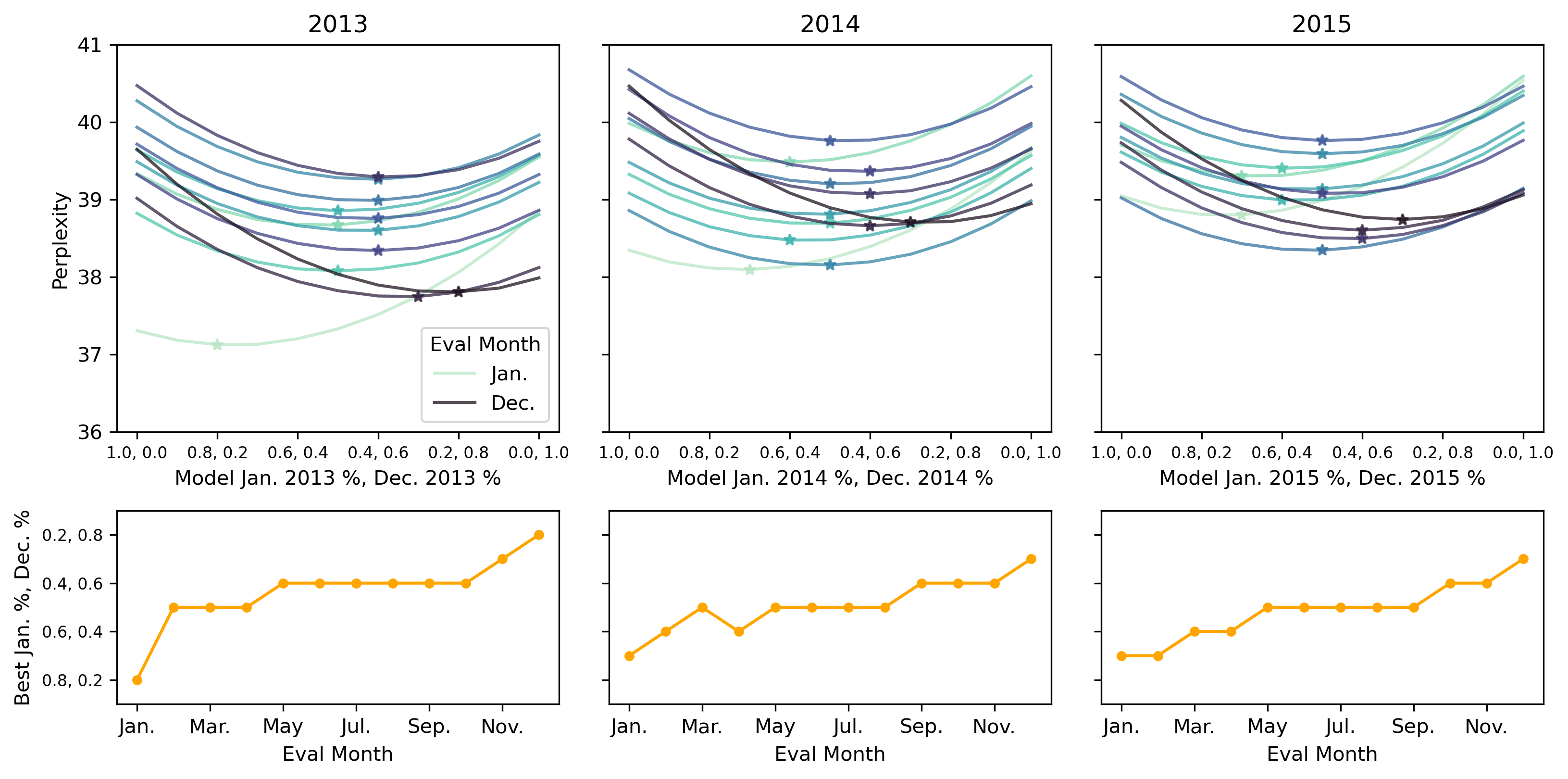}
    \caption{\textbf{Interpolating between two month vectors improves performance on the months between them.} We interpolate between January and December month vectors and evaluate on all other months within the same finetuning year. Like at the yearly scale, early months do better with a higher percentage of the January model and vice versa while middle months do best with a 50\% split between the models. The stars in the upper plots correspond to the best performing interpolations for each evaluation month; these optimums are mirrored in the lower line plots.}
    \label{fig:intervening_interpolation_months}
\end{figure*}

\begin{table}
    \resizebox{\columnwidth}{!}{%
    \begin{tabular}{l c c c}
        & \emph{Perplexity} ($\downarrow$) & \emph{Rouge} ($\uparrow$) & \emph{F1} ($\uparrow$) \\

        \toprule
        Method & WMT LM & NewsSum & PoliAff \\
        \midrule
        Start-year finetuned $(\tau_0)$ & 13.92 & 38.56 & 0.6886 \\
        End-year finetuned $(\tau_n)$ & 13.84 & 35.09 & 0.6967 \\
        $\frac{1}{2}(\tau_0 + \tau_n)$ & 13.77 & 38.86 & 0.7765 \\
        Best interpolations & 13.75 & 40.11 & 0.7941 \\ 
        Eval-year finetuned $(\tau_i)$ & \textbf{13.65} & \textbf{42.36} & \textbf{0.8341} \\
        \bottomrule
    \end{tabular}}
    \caption{\textbf{Interpolation between start and end-year finetuned models reduces temporal misalignment on intervening years.} T5-3b average performance on each year between start and end (non-inclusive). ``Best interpolations" use the best performing $\alpha$ values for each year.}
    \label{tab:intervening_interpolation_years}
\end{table}

\paragraph{Results} Figure \ref{fig:intervening_interpolation_years} shows that interpolating between start and end-year finetuned models improves performance on intervening years in both WMT LM and PoliAff tasks. Improvement is generally greatest on the exact middle years (2014 for WMT LM, 2017 for PoliAff) and decreases on years closer to start and end times. Patterns of improvement also vary depending on setting, with flatter changes in performance near $\alpha = 1.0, 0.0$ in PoliAff compared to WMT LM, and minimal improvements in NewsSum across $\alpha$s compared to the difference in performance between evaluation years. Table \ref{tab:intervening_interpolation_years} quantifies these changes, showing that interpolation closes the gap on intervening years between temporally aligned and misaligned models. Improvements are particularly large for PoliAff, nearly eight macro-F1 points just by averaging the start and end-year time vectors. 

Figure \ref{fig:intervening_interpolation_months} shows that these results extend to the monthly scale for WMT LM; we can interpolate between time vectors finetuned on January and December in a year to improve performance on the months between them. The best interpolations for each month follow an intuitive pattern, with a higher percentage of the January model leading to better performance on earlier months and vice versa. 
\subsection{Generalizing to the Future}
\label{subsec:task_analogy}

\begin{figure}
    \centering
    \includegraphics[width=\columnwidth]{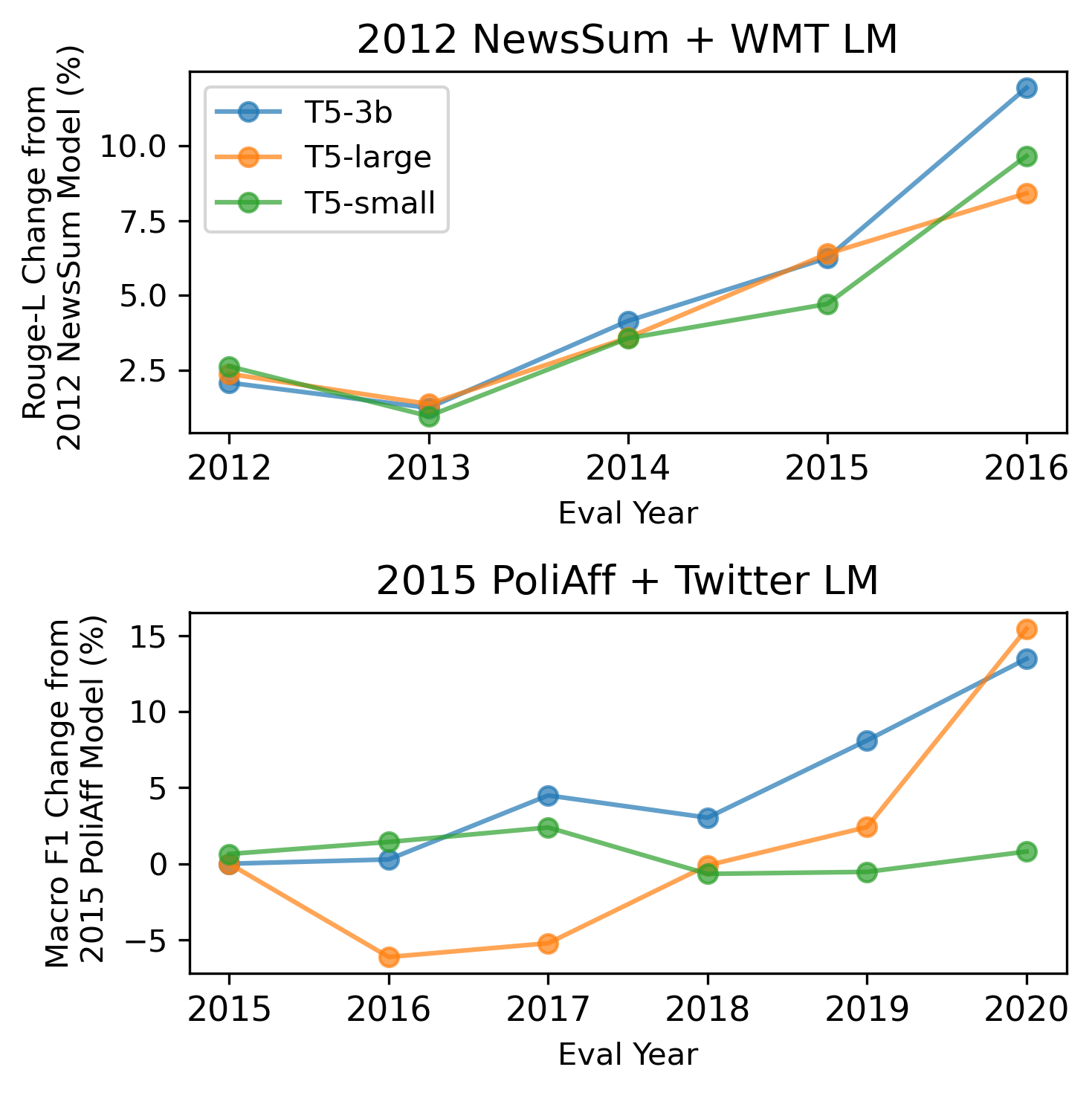}
    \caption{\textbf{Task analogies can offset downstream temporal misalignment without labeled data from the target time.} We report the performance of NewsSum and PoliAff T5 models updated using WMT LM and Twitter LM vectors for each target evaluation time. We report the percent improvement of the best updated model over 2012 NewsSum and 2015 PoliAff models on each target time for all model sizes.}
    \label{fig:task_analogy}
\end{figure}

The creation of labeled datasets lags behind corpera of raw text, which can be scraped automatically. As a result, language models that rely on supervision for finetuning are quickly outdated. Updating these models can be expensive, involving extra finetuning and creating labeled datasets from more recent examples. In this section, we present a new technique for updating task models finetuned on a source time period $j$ to a target time period $k$ with only unlabeled data from $j$, using task analogies \citep{ilharco2023editing}.

\paragraph{Method} Given language models with weights $\theta^{\text{LM}}_{j}, \theta^{\text{LM}}_{k}$ finetuned on unlabeled text from times $j, k$, and a task-specific model with weights $\theta_{j}$ finetuned on labeled data from time $j$, we perform the following arithmetic on the vectors:
\begin{align*}
\tau_j & = \theta_j - \theta_{pre}\\
\tau_j^{\text{LM}} & = \theta_j^{\text{LM}} - \theta_{pre} \\
\tau_k^{\text{LM}} & =  \theta_k^{\text{LM}} - \theta_{pre} \\
\tau_{k} & \approx \alpha_1 \cdot \tau_{j} + (\alpha_2 \cdot \tau^{\text{LM}}_k - \alpha_3 \cdot \tau^{\text{LM}}_j) \\
\theta_k  & = \tau_k + \theta_{pre}
\end{align*}

We evaluate our estimated $\theta_k$ on each target time $t_k$, sweeping over all combinations of $\alpha_1 \in [0.6, 0.8, \dots 2.2], \alpha_2, \alpha_3 \in [0.1, \dots 0.6]$ and reporting the best result compared to the original model $\theta_{j}$. In this section, we update a 2012 NewsSum model to 2013--2016, and a 2015 PoliAff model to 2016--2020 using WMT LM and Twitter LM time vectors respectively. 

\paragraph{Results} Task analogies improve performance on future years in both PoliAff and NewsSum tasks. Figure \ref{fig:task_analogy} shows that improvement compared to finetuning on the start year increases as the target and start years become more misaligned. Model size also affects performance, with T5-large and T5-3b showing greater improvements. In PoliAff, T5-small has no improvement over the baseline and T5-large task analogies perform worse than the baseline on 2016 and 2017 before improving on 2019 and 2020. Strangely, we find that only scaling $\alpha_1$ can also improve performance on future years. We report these $\alpha$ ablations and our results on two other classification tasks in Appendix \S\ref{subsec:time_vec_analogy_ablations}. We observe mostly similar results on these tasks, although there are task-specific inconsistencies.

\subsection{Generalizing to Multiple Time Periods}
\label{subsec:multi-year}

Because interpolations prove useful for generalizing to intervening and future time periods, we next test if we can build models that perform well on \textit{multiple} time periods by interpolating between all time vectors for a task. 

\paragraph{Method} 

We approach this problem with the \emph{model soup} technique \citep{wortsman2022model}. One of the key  practical advantages of soups is that constituent time-specific models can be trained independently (on smaller compute budgets) and combined at any time. Furthermore, the multi-year model does not need to be retrained to include new time periods; new time periods can be incorporated by merely growing the soup with additional finetuned models.

We attempt to create a multi-year model by following the recipe outlined by \citet{wortsman2022model}. They introduce two soup variants: the \emph{uniform soup} and \emph{greedy soup}. The uniform soup applies a uniform weight among all constituent models in the interpolation, while the greedy soup is an iterative procedure that only includes models in the soup that improves validation performance. We assess both variants here.

Our ``uniform time soup'' is $\theta_{\text{pre}} + \frac{1}{|T|} \sum_{t \in T} \tau_t$ where $T$ is the set of all years for a given task. For our ``greedy time soup,'' we implement a similar algorithm to \citet{wortsman2022model} which samples time vectors (with replacement) from each year in order of decreasing performance and adds them to the average model soup if they improve performance. 

To evaluate our ability to build models that generalize to multiple time periods, we measure the average performance across all evaluation years for each task. We compare our model soups against two baselines: 1) a model trained on all shuffled available data at once and 2) the best-performing model finetuned on only a single year of data. The all-year model is the most compute-intensive approach.

\paragraph{Results} Overwhelmingly, time soups perform worse than the model finetuned on all shuffled available data. For WMT LM and NewsSum, the uniform time soup performs worse than even the best single year model, despite having access to five times the amount of finetuning data. The greedy time soup only improves over the best single-year model on PoliAff with a single macro F1 point gain. These findings suggest that a model which generalizes to multiple time periods does not lie in a region of weight space bounded by models finetuned on single years of data. Future work may explore more sophisticated methods of merging which to induce better performing multi-year models. 

\subsection{Summary}

We propose methods for updating models to intervening, future, and multiple time periods using time vector arithmetic. We find that interpolating between two time vectors improves performance on unseen intervening times at both yearly and monthly scales. Similarly, we can improve performance on the future with unlabeled data from target times using time vector analogies. Building a multi-year model with a ``soup'' of time vectors, however, does not approach the performance of a model finetuned on all times at once. These results suggest that task arithmetic can be a simple way to update models to new times, but it does not help to improve genearlization across the board within a single model. 

\begin{table}
    \resizebox{\columnwidth}{!}{%
    \begin{tabular}{l c c c c c c c}
        & \emph{Perplexity } ($\downarrow$) & \emph{Rouge} ($\uparrow$) & \emph{F1} ($\uparrow$)\\
        \toprule

        Method & WMT LM & NewsSum & PoliAff \\
        \midrule
        Best single-year model & 34.45 & 38.95 & 0.7101 \\
        \midrule
        Uniform time soup & 34.70 & 33.05 & 0.6078 \\
        Greedy time soup & 34.45 & 38.95 & 0.7202 \\
        Training on all years & \textbf{29.17} & \textbf{40.07} & \textbf{0.7853} \\
        \bottomrule
    \end{tabular}}
    \caption{\textbf{Interpolation does not enable generalization to multiple time periods simultaneously.} Here, we measure  the average performance of models on all years. We compare multiple ways of building multi-year models; T5-small models finetuned to individual years or all years, and ``time soups'' created by averaging together all year time vectors for a task.}
    \label{table:single_vs_multi_year_models}
\end{table}

\section{Related Work}

\paragraph{Semantic Drift} Although changes in the full weight spaces of models over time have not been previously explored, semantic changes in word embeddings over time are well-documented \citep{hamilton2016diachronic}. Temporal misalignment \citep{bamler2017dynamic, gonen2021simple} and word analogies over time \citep{szymanski2017temporal} have also been studied in embeddings. Our work extends these analyses to the full set of language model parameters.

\paragraph{Temporal Misalignment} The phenomenon of temporal misalignment in language models has gained attention in the last three years. Moving from semantic drift to model misalignment in the twitter domain, \citet{jaidka2018diachronic} measure gender and age classifier degradation over time, \citet{rijhwani2020temporally} demonstrate the effect of temporal drift on named entity recognition, and \citet{loureiro2022timelms} find decay on a variety of language modeling and downstream tasks. \citet{lazaridou2021mind} extend these analyses to language modeling on News and Science domains and show that increasing model size does not help mitigate temporal misalignment. \citet{luu-etal-2022-time} compare temporal misalignment across a variety of downstream tasks, finding that degradation varies greatly over both domain and task. Using the same suite of tasks, \citet{longpre2023pretrainer} report similar degradation over time in pretraining regardless of model size. 

\paragraph{Updating LMs} Recent attempts at updating language models to new time periods have used a range of techniques. \citet{luu-etal-2022-time} find limited improvement with continued pretraining \citep{rottger2021temporal} on target times. Similar to the sequential updating setting, however, \citet{lazaridou2021mind} show that dynamic evaluation \cite{Gururangan2020DontSP} can improve language modeling performance on new times, but results in  forgetting the past. More recent techniques have been proposed for keeping models up to date in the QA domain by adding flags with the year for each example \citep{dhingra-etal-2022-time} or by discarding outdated facts \citep{zhang2023mitigating}. Unlike these methods, we consider the problem of updating models to new time periods without data in the target time and without additional finetuning. 

\paragraph{Interpolation} Our work draws heavily on recent techniques for editing models directly with interpolation and task analogies. Time vectors are an application of task vectors \citep{ilharco2023editing} to the time domain, our interpolation experiments are inspired by previous work on patching models for multiple tasks \citep{ilharco2022patching}, and our time soups are an application of models soups (averaging multiple models trained with different initializations) \citep{wortsman2022model}.
 
\section{Conclusion}

We connect studies of temporal misalignment and weight arithmetic with time vectors, formed by finetuning a model on a specific time period and then subtracting its pretrained weights. We show that the weights of time vectors are more similar if their corresponding times are closer and vice versa. These similarities are highly correlated to temporal misalignment at both yearly and monthly scales (which exhibit seasonal patterns). Leveraging this temporal structure in weight space, we induce new models that perform better on intervening years by interpolating between adjacent time vectors. Similarly, we use task analogies to improve downstream performance on future time periods using only unlabeled data from those times. These results show that task arithmetic can be a simple tool for combating temporal misalignment. 


\bibliography{custom}

\begin{thebibliography}{25}
\expandafter\ifx\csname natexlab\endcsname\relax\def\natexlab#1{#1}\fi

\bibitem[{Bamler and Mandt(2017)}]{bamler2017dynamic}
Robert Bamler and Stephan Mandt. 2017.
\newblock Dynamic word embeddings.
\newblock In \emph{International conference on Machine learning}, pages 380--389. PMLR.

\bibitem[{Barrault et~al.(2021)Barrault, Bojar, Bougares, Chatterjee, Costa-jussa, Federmann, Fishel, Fraser, Freitag, Graham, Grundkiewicz, Guzman, Haddow, Huck, Yepes, Koehn, Kocmi, Martins, Morishita, and Monz}]{wmt-2021-machine}
Loic Barrault, Ondrej Bojar, Fethi Bougares, Rajen Chatterjee, Marta~R. Costa-jussa, Christian Federmann, Mark Fishel, Alexander Fraser, Markus Freitag, Yvette Graham, Roman Grundkiewicz, Paco Guzman, Barry Haddow, Matthias Huck, Antonio~Jimeno Yepes, Philipp Koehn, Tom Kocmi, Andre Martins, Makoto Morishita, and Christof Monz, editors. 2021.
\newblock \href {https://aclanthology.org/2021.wmt-1.0} {\emph{Proceedings of the Sixth Conference on Machine Translation}}. Association for Computational Linguistics, Online.

\bibitem[{Dhingra et~al.(2022)Dhingra, Cole, Eisenschlos, Gillick, Eisenstein, and Cohen}]{dhingra-etal-2022-time}
Bhuwan Dhingra, Jeremy~R. Cole, Julian~Martin Eisenschlos, Daniel Gillick, Jacob Eisenstein, and William~W. Cohen. 2022.
\newblock \href {https://doi.org/10.1162/tacl_a_00459} {Time-aware language models as temporal knowledge bases}.
\newblock \emph{Transactions of the Association for Computational Linguistics}, 10:257--273.

\bibitem[{Gonen et~al.(2021)Gonen, Jawahar, Seddah, and Goldberg}]{gonen2021simple}
Hila Gonen, Ganesh Jawahar, Djam{\'e} Seddah, and Yoav Goldberg. 2021.
\newblock Simple, interpretable and stable method for detecting words with usage change across corpora.
\newblock \emph{arXiv preprint arXiv:2112.14330}.

\bibitem[{Grusky et~al.(2018)Grusky, Naaman, and Artzi}]{grusky2018newsroom}
Max Grusky, Mor Naaman, and Yoav Artzi. 2018.
\newblock Newsroom: A dataset of 1.3 million summaries with diverse extractive strategies.
\newblock \emph{arXiv preprint arXiv:1804.11283}.

\bibitem[{Gururangan et~al.(2020)Gururangan, Marasovi{\'c}, Swayamdipta, Lo, Beltagy, Downey, and Smith}]{Gururangan2020DontSP}
Suchin Gururangan, Ana Marasovi{\'c}, Swabha Swayamdipta, Kyle Lo, Iz~Beltagy, Doug Downey, and Noah~A. Smith. 2020.
\newblock \href {https://api.semanticscholar.org/CorpusID:216080466} {Don’t stop pretraining: Adapt language models to domains and tasks}.
\newblock \emph{ArXiv}, abs/2004.10964.

\bibitem[{Hamilton et~al.(2016)Hamilton, Leskovec, and Jurafsky}]{hamilton2016diachronic}
William~L Hamilton, Jure Leskovec, and Dan Jurafsky. 2016.
\newblock Diachronic word embeddings reveal statistical laws of semantic change.
\newblock \emph{arXiv preprint arXiv:1605.09096}.

\bibitem[{Hu et~al.(2021)Hu, Shen, Wallis, Allen-Zhu, Li, Wang, Wang, and Chen}]{hu2021lora}
Edward~J. Hu, Yelong Shen, Phillip Wallis, Zeyuan Allen-Zhu, Yuanzhi Li, Shean Wang, Lu~Wang, and Weizhu Chen. 2021.
\newblock \href {http://arxiv.org/abs/2106.09685} {Lora: Low-rank adaptation of large language models}.

\bibitem[{Ilharco et~al.(2023)Ilharco, Ribeiro, Wortsman, Gururangan, Schmidt, Hajishirzi, and Farhadi}]{ilharco2023editing}
Gabriel Ilharco, Marco~Tulio Ribeiro, Mitchell Wortsman, Suchin Gururangan, Ludwig Schmidt, Hannaneh Hajishirzi, and Ali Farhadi. 2023.
\newblock \href {http://arxiv.org/abs/2212.04089} {Editing models with task arithmetic}.

\bibitem[{Ilharco et~al.(2022)Ilharco, Wortsman, Gadre, Song, Hajishirzi, Kornblith, Farhadi, and Schmidt}]{ilharco2022patching}
Gabriel Ilharco, Mitchell Wortsman, Samir~Yitzhak Gadre, Shuran Song, Hannaneh Hajishirzi, Simon Kornblith, Ali Farhadi, and Ludwig Schmidt. 2022.
\newblock Patching open-vocabulary models by interpolating weights.
\newblock \emph{Advances in Neural Information Processing Systems}, 35:29262--29277.

\bibitem[{Jaidka et~al.(2018)Jaidka, Chhaya, and Ungar}]{jaidka2018diachronic}
Kokil Jaidka, Niyati Chhaya, and Lyle Ungar. 2018.
\newblock Diachronic degradation of language models: Insights from social media.
\newblock In \emph{Proceedings of the 56th Annual Meeting of the Association for Computational Linguistics (Volume 2: Short Papers)}, pages 195--200.

\bibitem[{Lazaridou et~al.(2021)Lazaridou, Kuncoro, Gribovskaya, Agrawal, Liska, Terzi, Gimenez, de~Masson~d'Autume, Kocisky, Ruder, Yogatama, Cao, Young, and Blunsom}]{lazaridou2021mind}
Angeliki Lazaridou, Adhiguna Kuncoro, Elena Gribovskaya, Devang Agrawal, Adam Liska, Tayfun Terzi, Mai Gimenez, Cyprien de~Masson~d'Autume, Tomas Kocisky, Sebastian Ruder, Dani Yogatama, Kris Cao, Susannah Young, and Phil Blunsom. 2021.
\newblock \href {http://arxiv.org/abs/2102.01951} {Mind the gap: Assessing temporal generalization in neural language models}.

\bibitem[{Lester et~al.(2021)Lester, Al-Rfou, and Constant}]{lester2021power}
Brian Lester, Rami Al-Rfou, and Noah Constant. 2021.
\newblock \href {http://arxiv.org/abs/2104.08691} {The power of scale for parameter-efficient prompt tuning}.

\bibitem[{Li et~al.(2022)Li, Gururangan, Dettmers, Lewis, Althoff, Smith, and Zettlemoyer}]{li2022branchtrainmerge}
Margaret Li, Suchin Gururangan, Tim Dettmers, Mike Lewis, Tim Althoff, Noah~A. Smith, and Luke Zettlemoyer. 2022.
\newblock \href {http://arxiv.org/abs/2208.03306} {Branch-train-merge: Embarrassingly parallel training of expert language models}.

\bibitem[{Longpre et~al.(2023)Longpre, Yauney, Reif, Lee, Roberts, Zoph, Zhou, Wei, Robinson, Mimno et~al.}]{longpre2023pretrainer}
Shayne Longpre, Gregory Yauney, Emily Reif, Katherine Lee, Adam Roberts, Barret Zoph, Denny Zhou, Jason Wei, Kevin Robinson, David Mimno, et~al. 2023.
\newblock A pretrainer's guide to training data: Measuring the effects of data age, domain coverage, quality, \& toxicity.
\newblock \emph{arXiv preprint arXiv:2305.13169}.

\bibitem[{Loureiro et~al.(2022)Loureiro, Barbieri, Neves, Anke, and Camacho-Collados}]{loureiro2022timelms}
Daniel Loureiro, Francesco Barbieri, Leonardo Neves, Luis~Espinosa Anke, and Jose Camacho-Collados. 2022.
\newblock Timelms: Diachronic language models from twitter.
\newblock \emph{arXiv preprint arXiv:2202.03829}.

\bibitem[{Luu et~al.(2022)Luu, Khashabi, Gururangan, Mandyam, and Smith}]{luu-etal-2022-time}
Kelvin Luu, Daniel Khashabi, Suchin Gururangan, Karishma Mandyam, and Noah~A. Smith. 2022.
\newblock \href {https://doi.org/10.18653/v1/2022.naacl-main.435} {Time waits for no one! analysis and challenges of temporal misalignment}.
\newblock In \emph{Proceedings of the 2022 Conference of the North American Chapter of the Association for Computational Linguistics: Human Language Technologies}, pages 5944--5958, Seattle, United States. Association for Computational Linguistics.

\bibitem[{Ortiz-Jim{\'e}nez et~al.(2023)Ortiz-Jim{\'e}nez, Favero, and Frossard}]{OrtizJimnez2023TaskAI}
Guillermo Ortiz-Jim{\'e}nez, Alessandro Favero, and Pascal Frossard. 2023.
\newblock \href {https://api.semanticscholar.org/CorpusID:258832777} {Task arithmetic in the tangent space: Improved editing of pre-trained models}.
\newblock \emph{ArXiv}, abs/2305.12827.

\bibitem[{Raffel et~al.(2023)Raffel, Shazeer, Roberts, Lee, Narang, Matena, Zhou, Li, and Liu}]{raffel2023exploring}
Colin Raffel, Noam Shazeer, Adam Roberts, Katherine Lee, Sharan Narang, Michael Matena, Yanqi Zhou, Wei Li, and Peter~J. Liu. 2023.
\newblock \href {http://arxiv.org/abs/1910.10683} {Exploring the limits of transfer learning with a unified text-to-text transformer}.

\bibitem[{Rijhwani and Preo{\c{t}}iuc-Pietro(2020)}]{rijhwani2020temporally}
Shruti Rijhwani and Daniel Preo{\c{t}}iuc-Pietro. 2020.
\newblock Temporally-informed analysis of named entity recognition.
\newblock In \emph{Proceedings of the 58th Annual Meeting of the Association for Computational Linguistics}, pages 7605--7617.

\bibitem[{R{\"o}ttger and Pierrehumbert(2021)}]{rottger2021temporal}
Paul R{\"o}ttger and Janet~B Pierrehumbert. 2021.
\newblock Temporal adaptation of bert and performance on downstream document classification: Insights from social media.
\newblock \emph{arXiv preprint arXiv:2104.08116}.

\bibitem[{Szymanski(2017)}]{szymanski2017temporal}
Terrence Szymanski. 2017.
\newblock Temporal word analogies: Identifying lexical replacement with diachronic word embeddings.
\newblock In \emph{Proceedings of the 55th annual meeting of the association for computational linguistics (volume 2: short papers)}, pages 448--453.

\bibitem[{Wortsman et~al.(2022)Wortsman, Ilharco, Gadre, Roelofs, Gontijo-Lopes, Morcos, Namkoong, Farhadi, Carmon, Kornblith et~al.}]{wortsman2022model}
Mitchell Wortsman, Gabriel Ilharco, Samir~Ya Gadre, Rebecca Roelofs, Raphael Gontijo-Lopes, Ari~S Morcos, Hongseok Namkoong, Ali Farhadi, Yair Carmon, Simon Kornblith, et~al. 2022.
\newblock Model soups: averaging weights of multiple fine-tuned models improves accuracy without increasing inference time.
\newblock In \emph{International Conference on Machine Learning}, pages 23965--23998. PMLR.

\bibitem[{Wortsman et~al.(2021)Wortsman, Ilharco, Li, Kim, Hajishirzi, Farhadi, Namkoong, and Schmidt}]{Wortsman2021RobustFO}
Mitchell Wortsman, Gabriel Ilharco, Mike Li, Jong~Wook Kim, Hannaneh Hajishirzi, Ali Farhadi, Hongseok Namkoong, and Ludwig Schmidt. 2021.
\newblock \href {https://api.semanticscholar.org/CorpusID:237420687} {Robust fine-tuning of zero-shot models}.
\newblock \emph{2022 IEEE/CVF Conference on Computer Vision and Pattern Recognition (CVPR)}, pages 7949--7961.

\bibitem[{Zhang and Choi(2023)}]{zhang2023mitigating}
Michael~JQ Zhang and Eunsol Choi. 2023.
\newblock Mitigating temporal misalignment by discarding outdated facts.
\newblock \emph{arXiv preprint arXiv:2305.14824}.

\end{thebibliography}
\bibliographystyle{acl_natbib}

\appendix

\section{Appendix}
\label{sec:appendix}

\subsection{Yearly Misalignment with Other Tasks and T5 Sizes}
\label{subsec:other_yearly_misalignment}

In this section, we report raw performance degradation over time on four downstream and three language modeling tasks with three sizes of T5. We evaluate on all tasks in the main paper plus Newsroom Source Classification (NewsCls) and AI Venue Classification (AIC) from \citet{luu-etal-2022-time}. We also create a third science domain language modeling task from abstracts in the Kaggle arXiv dataset\footnote{https://www.kaggle.com/datasets/Cornell-University/arxiv/data}. For each group of three years from 2006-2008 to 2018-2020 we randomly sample 26-38M and 2.6-3.9M BPE tokens (150MB and 15MB of text) of arXiv paper abstracts for train and test splits respectively.

\begin{figure*}
    \centering
    \includegraphics[width=\textwidth]{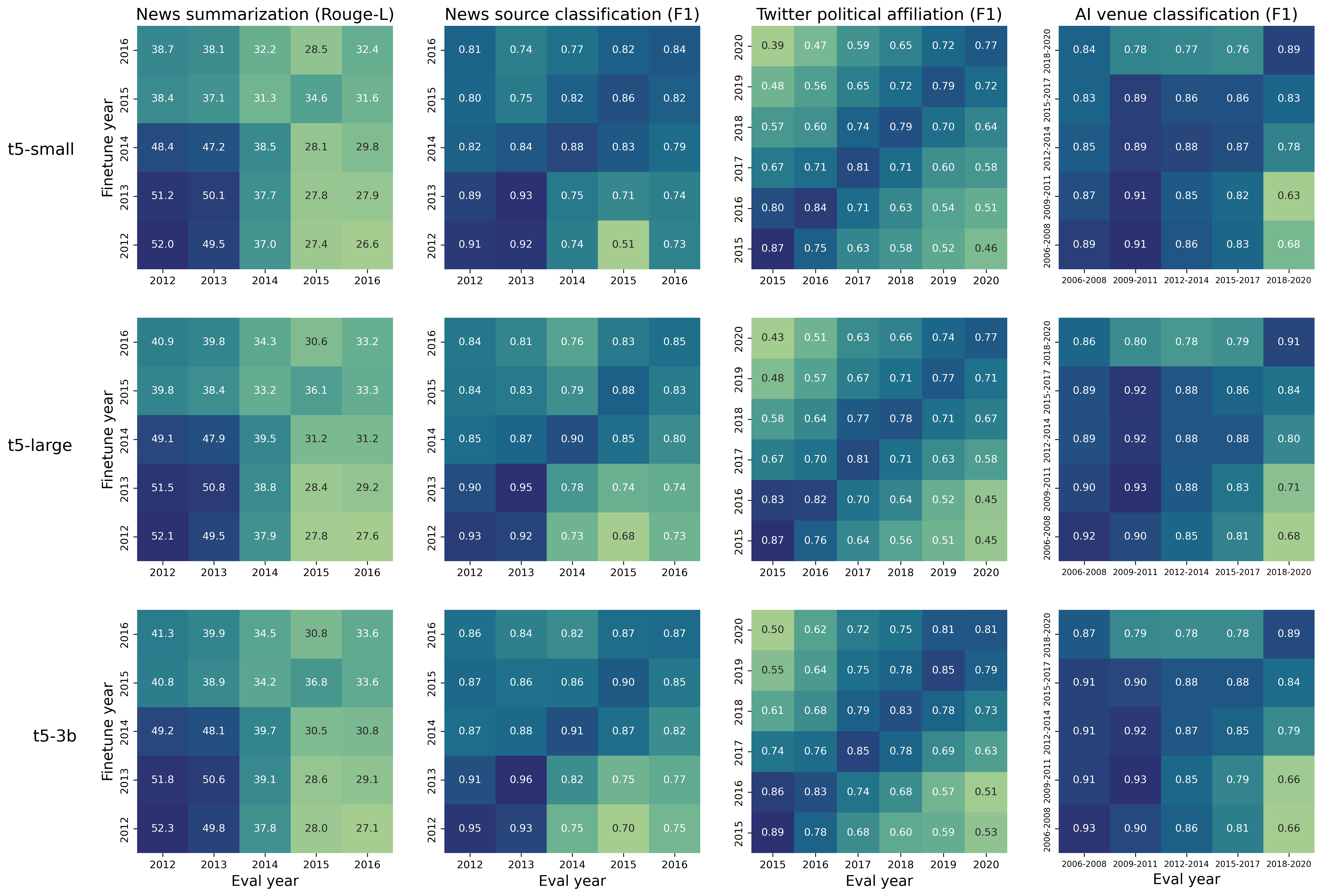}
    \caption{Yearly downstream performance degradation on four tasks and three T5 sizes.}
    \label{fig:yearly_downstream_decay_big}
\end{figure*}

\begin{figure*}
    \centering
    \includegraphics[width=0.75\textwidth]{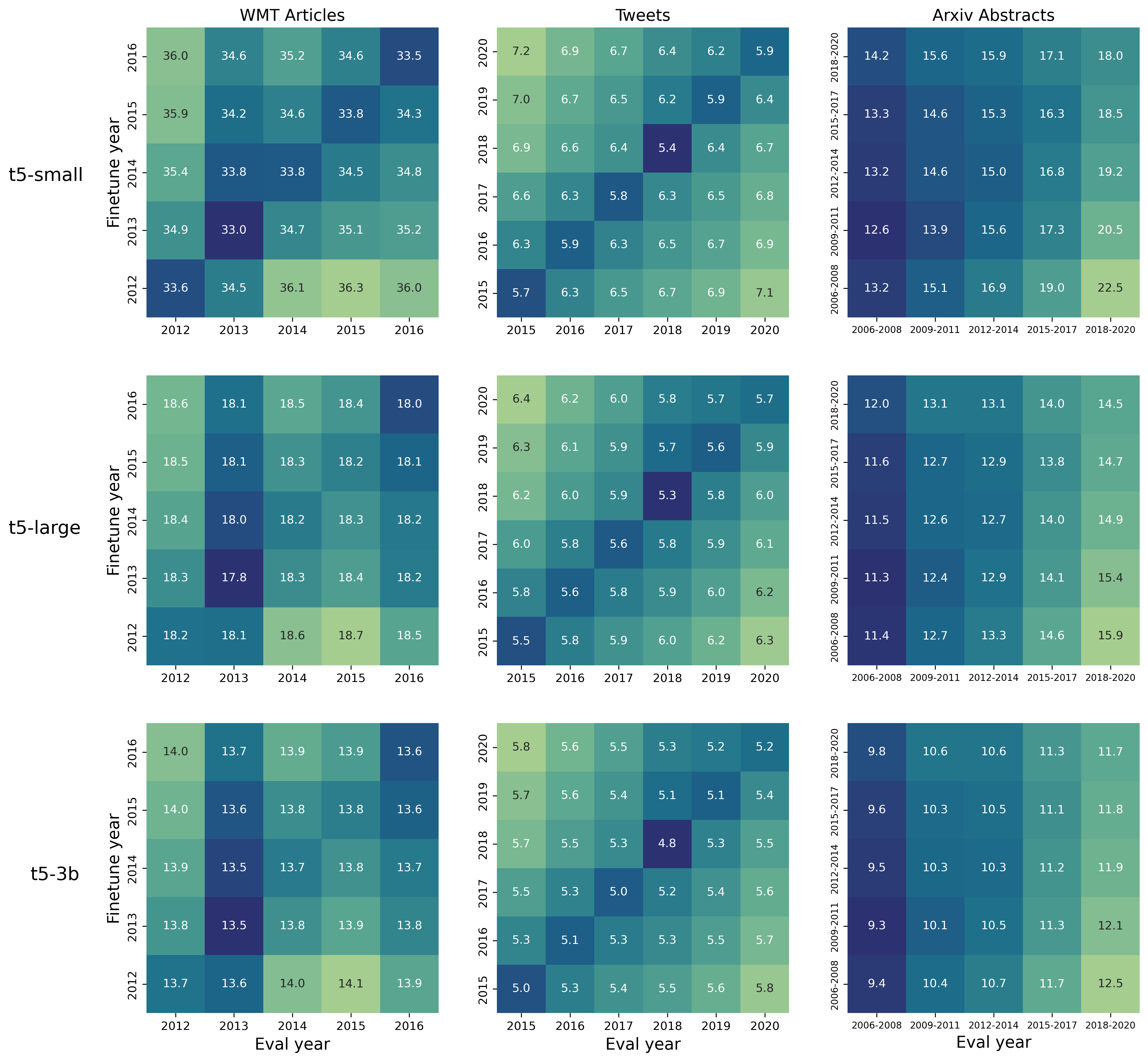}
    \caption{Yearly language modeling perplexity decay on three tasks and three T5 sizes.}
    \label{fig:yearly_lm_decay_big}
\end{figure*}

Figures \ref{fig:yearly_downstream_decay_big} and \ref{fig:yearly_lm_decay_big} are yearly degradation heatmaps for each model size and task. These results show that normalizing performance by the average on each evaluation time helps account for variations in test splits. ArXiv language modeling and NewsSum, for example, have large differences in performance on evaluation years regardless of finetuning year.

\begin{figure}
    \centering
    \includegraphics[width=\columnwidth]{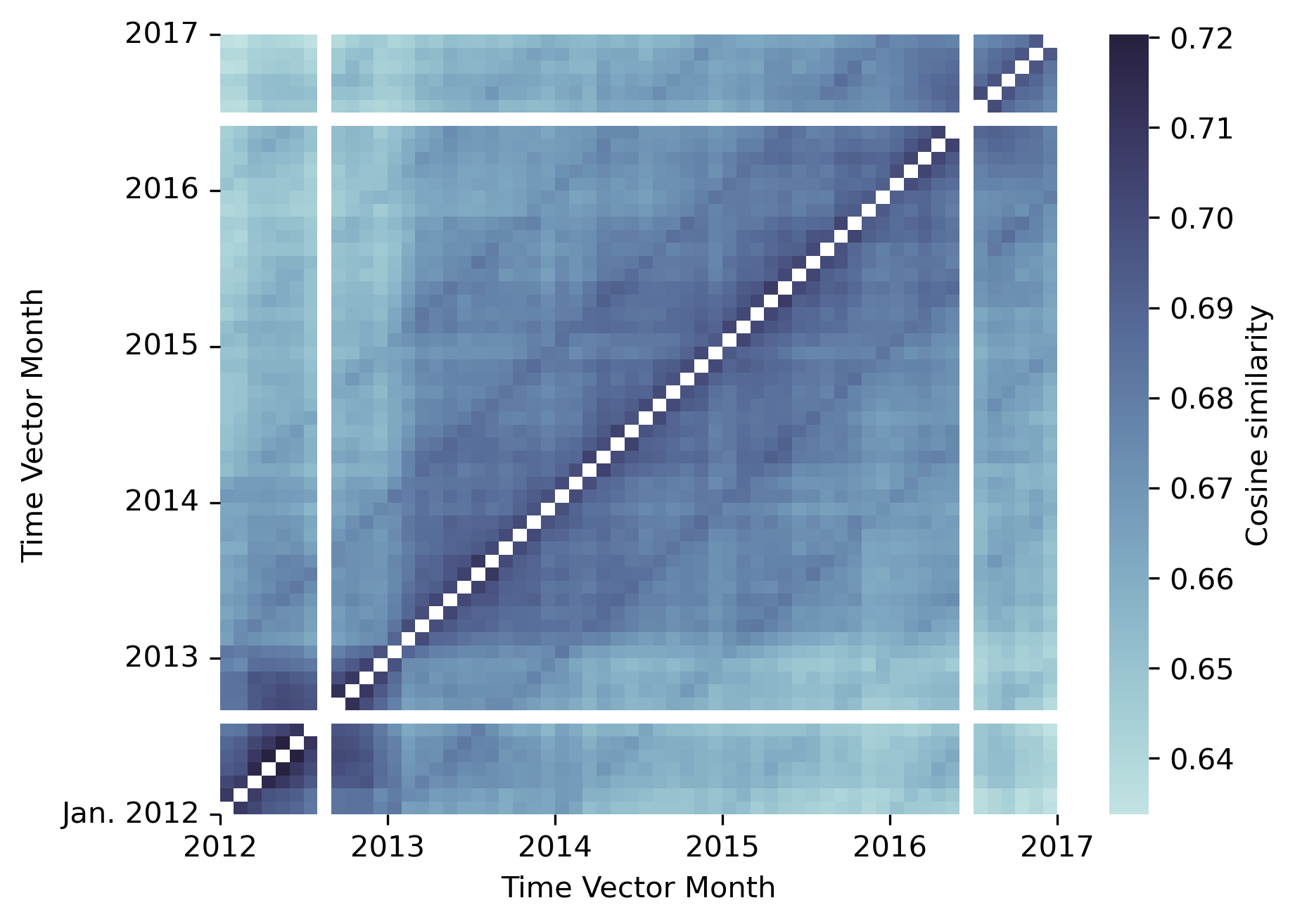}
    \caption{\textbf{Cosine similarity between monthly time vectors also exhibits seasonality.} We observe similar "stripes" every 12 months when measuring the cosine similarity between each pair of T5-small WMT month vectors. The correlation between this heatmap (including the diagonal) and figure \ref{fig:monthly_misalignment} is $-0.667$ with $p < 1\times 10^{-16}$.}
    \label{fig:monthly_cos_sims}
\end{figure}

\begin{figure*}
    \centering
    \includegraphics[width=\textwidth]{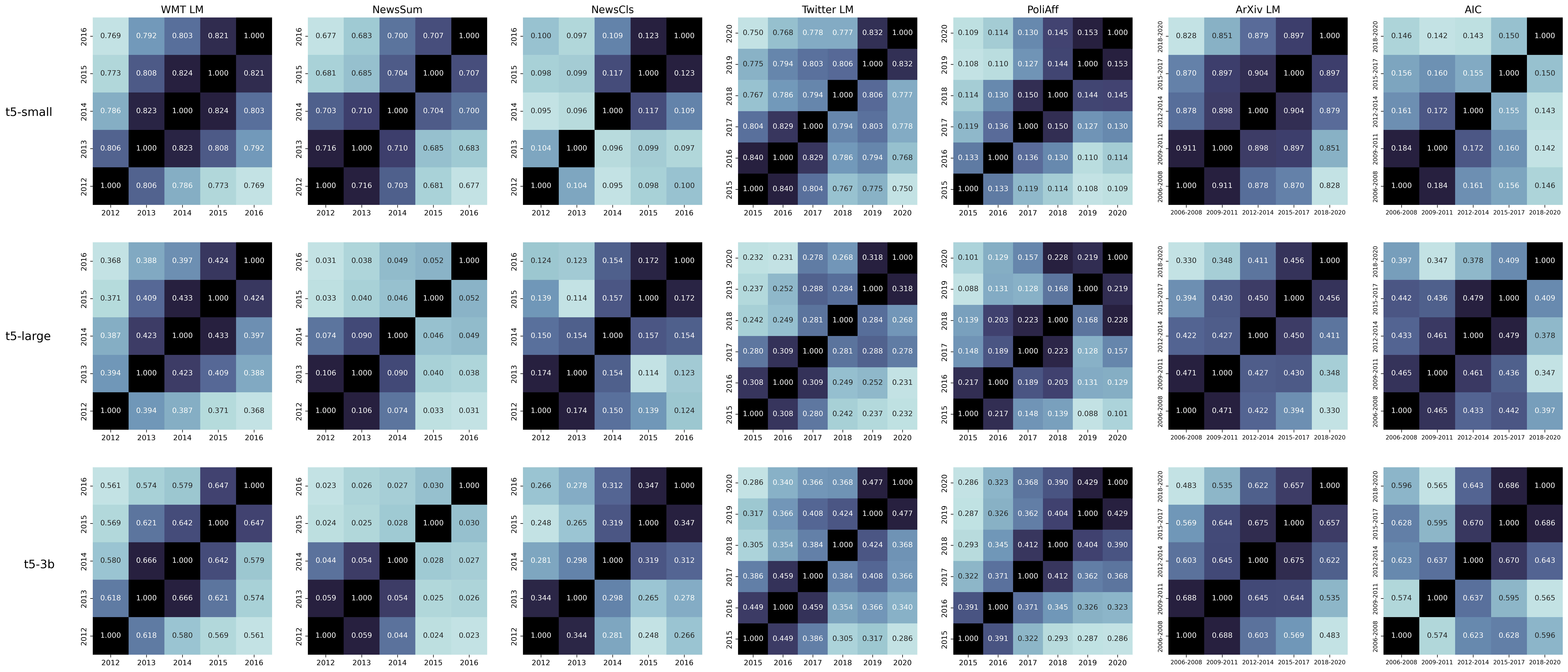}
    \caption{Cosine similarities between all pairs of year time vectors for all tasks and model sizes.}
    \label{fig:yearly_cos_sim_heatmaps}
\end{figure*}

\subsection{Task Variations in Linear Yearly Degradation}
\label{subsec:linear_deg_variations}

Like \citet{luu-etal-2022-time}, we find differences across domain and task in the rate and linearity of year-to-year decay. TD scores measure the average rate of performance degredation for each year of misalignment between train and test time periods \citep{luu-etal-2022-time}. We find the rate of decay using a linear least squares regression and average rates for each task over all evaluations. Table \ref{table:td_scores} shows TD scores \citep{luu-etal-2022-time} for all tasks and T5-sizes. We also compare TD scores calculated from raw performance to TD scores calculated from performance normalized by the average on each evaluation year. In general, percent performance difference from the mean on an evaluation year decays more linearly than raw performance.

\begin{table*}
    \resizebox{\textwidth}{!}{%
    \begin{tabular}{c | l c c c c c c c}
        \toprule
        Normalized? & T5 Size & WMT LM & NewsSum & NewsCls & Twitter LM & PoliAff & ArXiv LM & AIC \\
        \midrule
         & small & -0.67 (0.81) & 2.21 (0.51) & 0.05 (0.67) & -0.35 (0.97) & 0.08 (0.98) & -0.59 (0.65) & 0.03 (0.55) \\
        No & large & -0.10 (0.34) & 2.07 (0.53) & 0.04 (0.61) & -0.20 (0.97) & 0.07 (0.97) & -0.20 (0.67) & 0.03 (0.50) \\
         & 3b & -0.07 (0.34) & 2.12 (0.53) & 0.04 (0.67) & -0.20 (0.97) & 0.07 (0.95) & -0.13 (0.66) & 0.03 (0.40) \\
        \midrule
         & small & -1.70 (0.90) & 6.99 (0.87) & 6.43 (0.74) & -4.52 (0.89) & 10.47 (0.95) & -2.61 (0.94) & 2.93 (0.57) \\
        Yes & large & -0.56 (0.92) & 6.27 (0.89) & 5.33 (0.84) & -2.64 (0.91) & 9.57 (0.94) & -1.24 (0.93) & 2.53 (0.51) \\
         & 3b & -0.52 (0.93) & 6.44 (0.88) & 4.72 (0.84) & -2.90 (0.91) & 7.66 (0.91) & -0.96 (0.94) & 3.12 (0.61) \\
        \bottomrule
    \end{tabular}}
    \caption{TD scores for all tasks and T5 sizes for raw performance and performance divide by the average on each eval. year. Variance explained by the TD score linear fit in parentheses. TD scores calculated with normalized performance decay have generally higher $R^2$ scores, except on Twitter LM and PoliAff, and are easier to compare.}
    \label{table:td_scores}
\end{table*}

\subsection{Yearly and Monthly Cosine Similarities}
In this section, we report cosine similarity between each pair of yearly and monthly time vectors. Figure \ref{fig:yearly_cos_sim_heatmaps} shows cosine similarity between every pair of year vectors for each T5-size and task. Figure \ref{fig:monthly_cos_sims} shows cosine similarity between each pair of T5-small monthly WMT LM time vectors. Similar to performance, year-to-year degradation in cosine similarity between task vectors appears to be linear regardless of setting. Like figure \ref{fig:monthly_misalignment}, we observe seasonal "stripes" every 12 months from the diagonal \ref{fig:monthly_cos_sims}.

\begin{figure}
    \centering
    \includegraphics[width=\columnwidth]{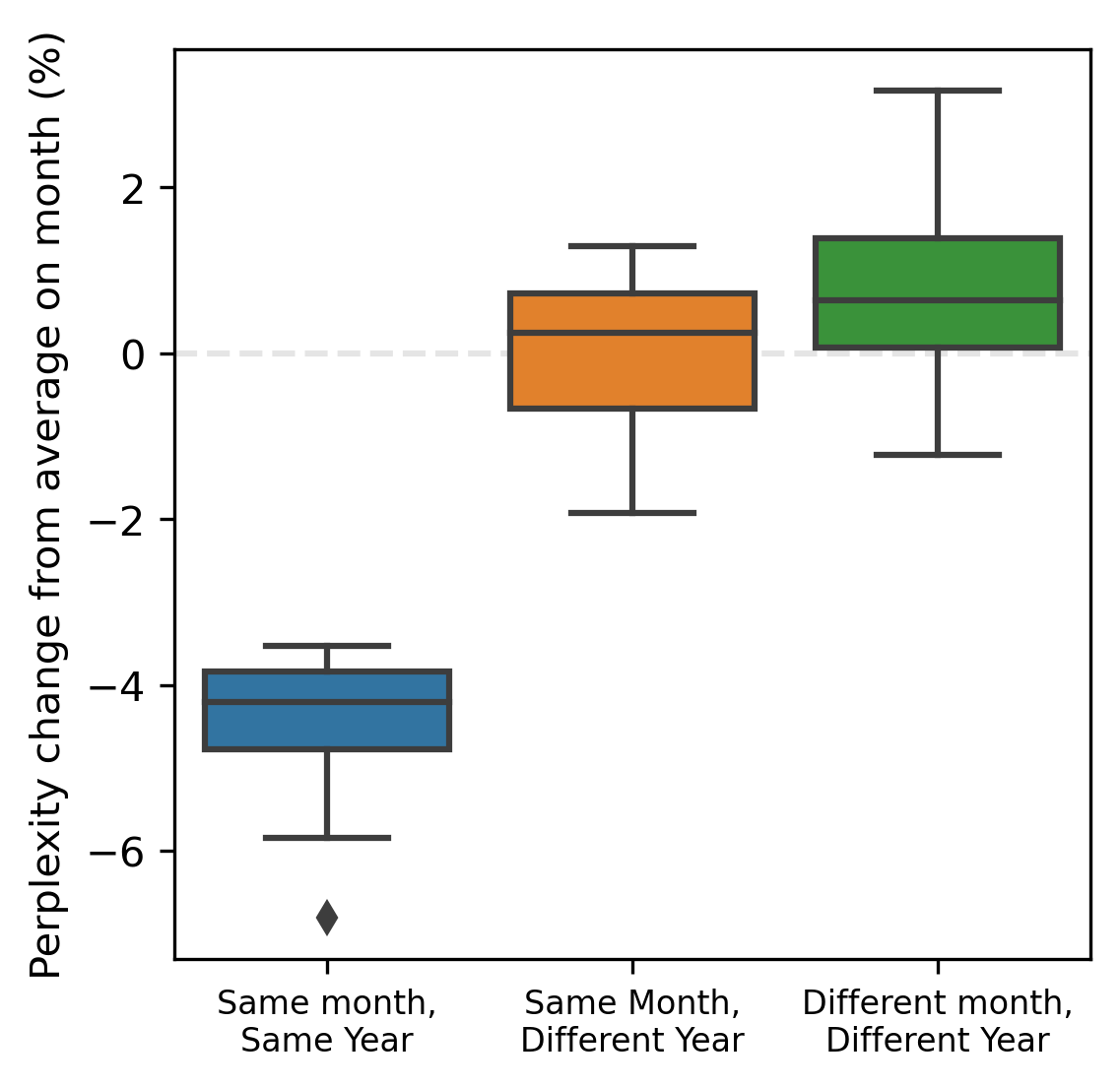}
    \caption{\textbf{Seasonality makes a small, but noticeable impact on monthly misalignment.} Distribution of perplexity change from the mean for aligned finetuning and evaluation months (left, mean=-4.36), seasonal "stripes" (middle, mean=0.04), and all finetuning and evaluation combinations which share neither the same month nor year (right, mean=0.77).}
    \label{fig:monthly-violins}
\end{figure}

\subsection{Temporal Degradation in Online Settings}
\label{subsec:online}

Our work so far illustrates temporal misalignment on static time splits. However, in practice, we usually deploy language models in online settings, meaning that they are continually updated with the latest data, and we do not have access to data from all training years simultaneously. 

\begin{figure*}
    \centering
    \includegraphics[width=0.8\textwidth]{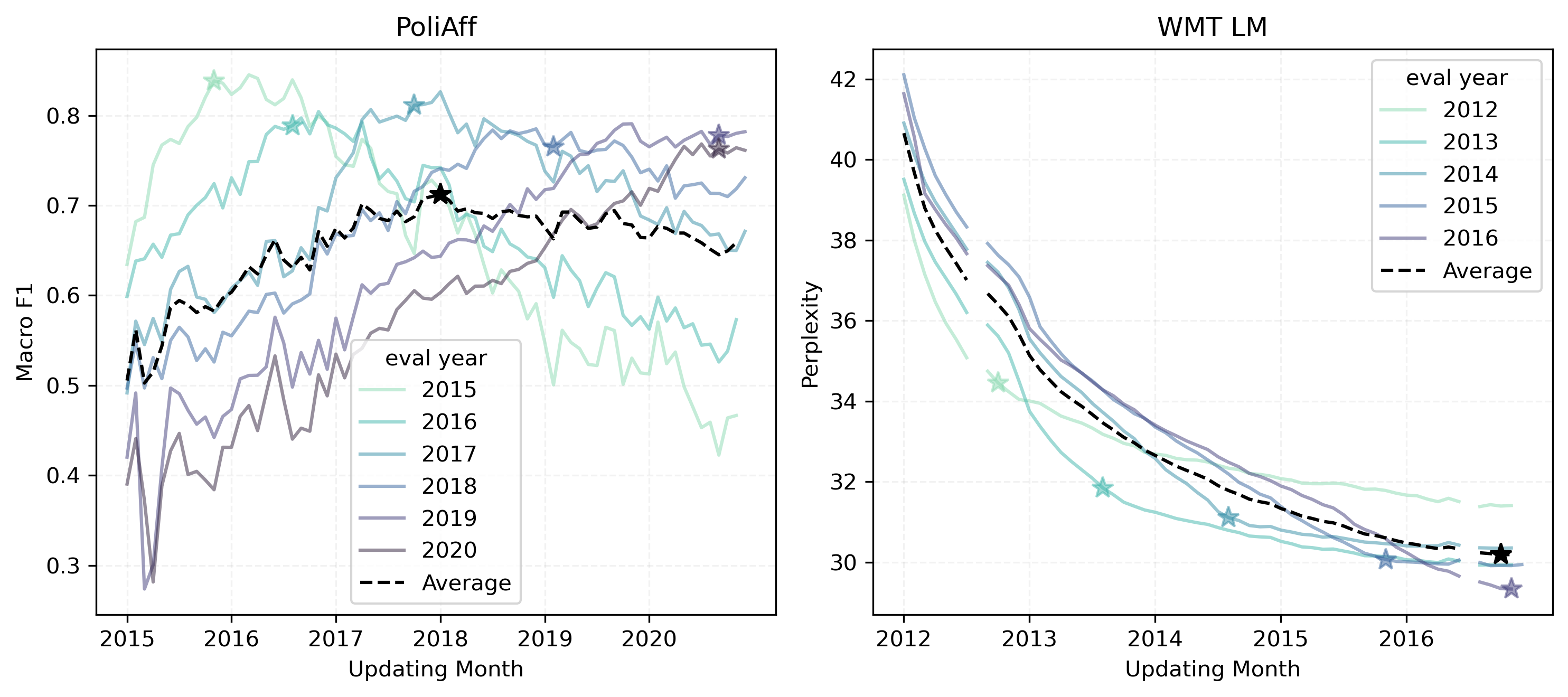}
    \caption{\textbf{In online settings, language model performance degrades on earlier time periods.} We show macro F1 and perplexity on each year split of PoliAff and WMT LM respectively after sequentially finetuning T5-small on each new month of task data. PoliAff performance over all years plateaus after finetuning on months up to 2018. WMT performance continues to improve with more data, but perplexity decrease slows on earlier years. Starred points are where performance on a year is best relative to the average performance on all years.}
    \label{fig:online_perf}
\end{figure*}

To show how temporal misalignment manifests in these settings, we first sort all the training data from the PoliAff and WMT tasks by month, and finetune T5-small on each task separately. We display the performance of the LM on every year throughout training in Figure \ref{fig:online_perf}. As expected, for PoliAff, we see that the performance of models on a particular year peak at the final month of that year, and then gradually degrade as the model continues training. 

For language modeling on WMT data, performance consistently \emph{improves} during training, regardless of the evaluation year. However, perplexity reduces more slowly in earlier years as we continue training. These results suggest that temporal misalignment may manifest differently in online settings based on the training setup and task.

\subsection{Online Cosine Similarities}
\label{subsec:cos_sims_online}

We study the relationship between performance degradation and cosine similarity during \emph{online training}. Recall that in the online setting, we perform a single finetuning run on the Poliaff and WMT tasks (after ordering their training data by month), and measure performance on each year throughout training. To study how time vectors move throughout space in this setting, we measure the cosine similarity between the time vector of the model trained up to month $m$ and each yearly time vector for the PoliAff and WMT tasks.

We find that the cosine similarity to each time vector \textit{decreases} as the online model is updated past the first 12 months of data. This means that online models' peak similarity to earlier years tends to be higher than those to later years since the they make up a smaller part of its total finetuning set. Like our experiments with soups of time vectors in section \S\ref{subsec:multi-year}, this indicates that models trained on multiple years of data lay outside a region defined by single-year models. 

To account for these decreases, we normalize the similarity to each year time vector by its average after updating on all months in Figure \ref{fig:online_perf}. Our results reveal that the vector for our online model is relatively most similar to each year vector after finetuning on the months in that year. 

\begin{figure*}
    \centering
    \includegraphics[width=0.8\textwidth]{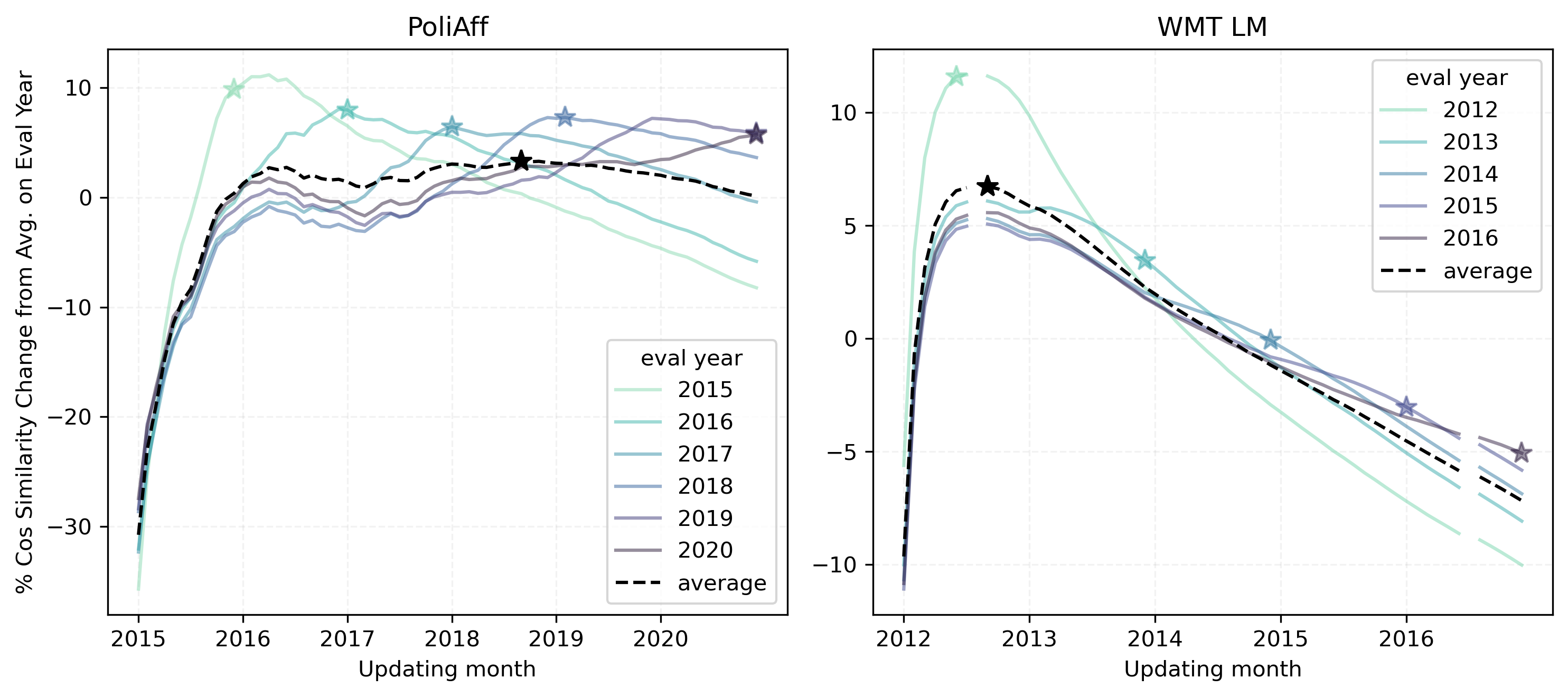}
    \caption{\textbf{Cosine similarity between an online time vector and a year vector peaks relative to other years after updating on data for that year.} We show cosine similarity between each monthly checkpoint of online T5-small time vectors and yearly vectors for PoliAff and WMT LM. To account for overall decreases in similarity as online time vectors are updated, we normalize similarities to each year vector by the mean similarity to that year over all checkpoints. We star the point for each year vector where its cosine similarity to the online model is largest relative to the average on all years.}
    \label{fig:online_perf_cos_sims}
\end{figure*}

\subsection{Time Vector Analogy Ablations}
\label{subsec:time_vec_analogy_ablations}

In this section, we ablate our time vector analogy experiment to determine the effects of only adding the LM vector from the target time, and only scaling the weights of the initial time vector. For $\tau_{k} \approx \alpha_1 \cdot \tau_{j} + (\alpha_2 \cdot \tau^{\text{LM}}_k - \alpha_3 \cdot \tau^{\text{LM}}_j)$, we define our "task addition" ablation for $\alpha_3 = 0, \alpha_1, \alpha_2 \ne 0$, and our "scaling only" ablation for $\alpha_1 \ne 0, \alpha_2, \alpha_3 = 0$

\begin{figure*}
    \centering
    \includegraphics[width=\textwidth]{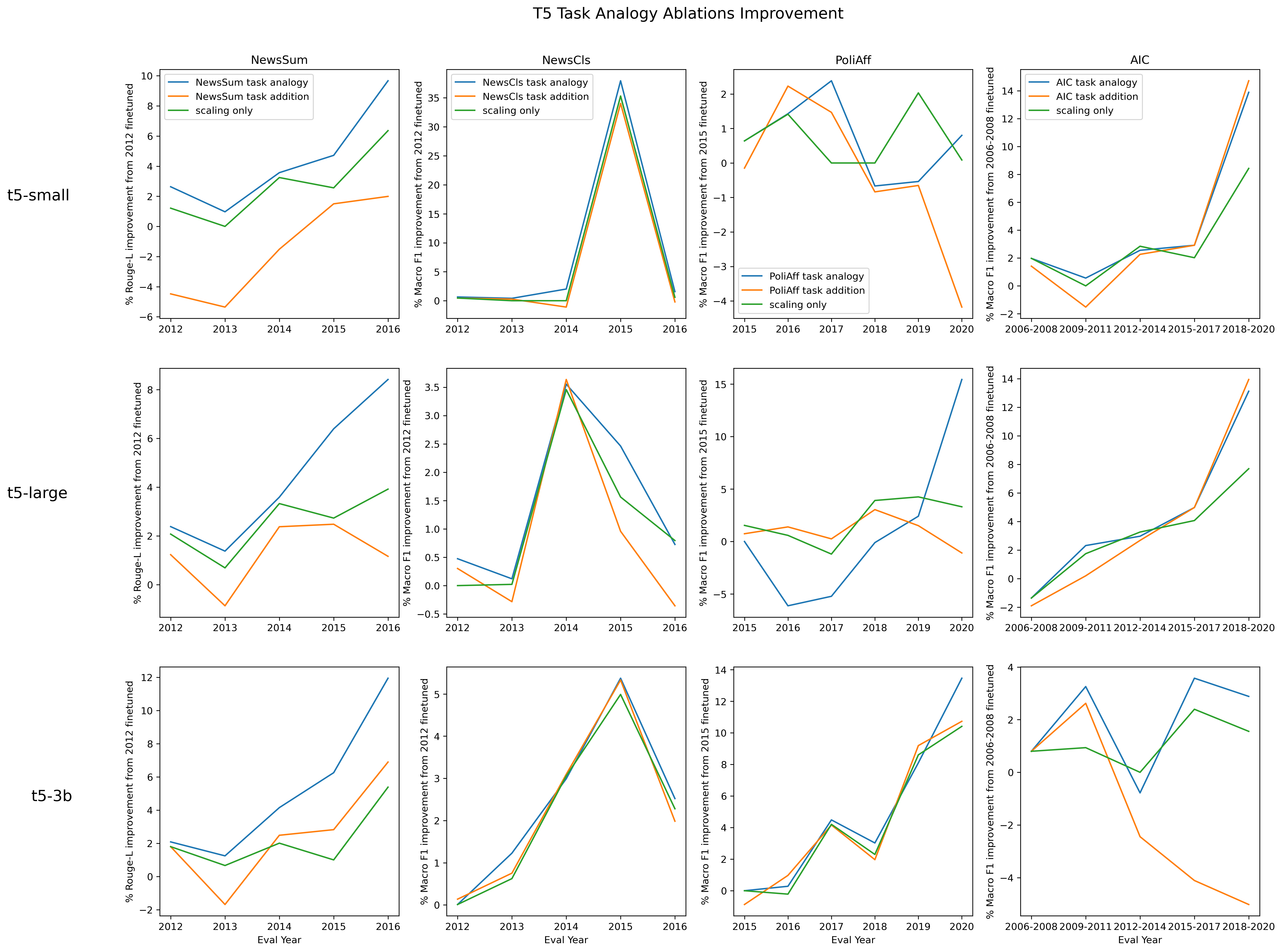}
    \caption{Time vector analogy ablations for three sizes of T5. Given the time vector analogy $\tau_{k} \approx \alpha_1 \cdot \tau_{j} + (\alpha_2 \cdot \tau^{\text{LM}}_k - \alpha_3 \cdot \tau^{\text{LM}}_j), \alpha_1, \alpha_2, \alpha_3 \ne 0$, we define "task addition" to be only adding the language modeling vector (i.e $\alpha_1, \alpha_2 \ne 0, \alpha_3 = 0$), and "scaling only" to be only scaling the base $\tau_j$ model (i.e $\alpha_1 \ne 0, \alpha_2, \alpha_3 = 0$). We sweep over the same alpha combinations as in \S\ref{subsec:task_analogy} and report the best results for each target year, task, and T5-size.}
    \label{fig:time_vec_analogy_ablations}
\end{figure*}

We report the best results after sweeping over the same $\alpha$ ranges from \S\ref{subsec:task_analogy} with the added constraints in figure \ref{fig:time_vec_analogy_ablations}. While task analogies generally perform best across tasks and T5-sizes (especially as $\tau_j$ and $\tau_k$ become more misaligned), we find that ablating $\tau^{\text{LM}}_k$ and $\tau^{\text{LM}}_j$ can still improve over the base $\tau_j$ model. Surprisingly, \textit{only scaling} $\tau_j$ also improves over the initial model on many tasks.

\subsection{Temporal Misalignment Affects Some Parameters More than Others}
In this section, we explore whether we can reduce temporal misalignment by swapping parameter weights from a model trained on a misaligned year with those of the model trained on the target year. For example, we substitute the QKV attention layers from a model finetuned on 2015 PoliAff with those finetuned on 2020 PoliAff and evaluate on 2020 data. In table \ref{tab:swapping} we evaluate the start-year finetuned models for each task on the end times (e.g. start = 2012 for WMT LM, end = 2016) with various parameter weights swapped with the end-year finetuned model.

From these experiments, we find that we can improve performance on a target time by swapping out weights with a time vector finetuned on that time. Surprisingly, swapping embeddings with the target time vector makes very little difference, except in language modeling tasks, and swapping all non-embedding weights with a target time almost reaches the performance the target time-specific models for downstream tasks. Swapping only feed-forward or attention layers also improves performance on the target time, suggesting temporal misalignment is somewhat isolated to those model regions in downstream tasks. \\

\begin{table*}
\begin{tabular}{l c c c c c c c}
    \toprule
    Swapped Params & WMT LM & NewsSum & NewsCls & Twitter LM & PoliAff & ArXiv LM & AIC \\
    \midrule
    \textit{None} & \textit{35.72} & \textit{35.11} & \textit{0.7232} & \textit{6.69} & \textit{0.5903} & \textit{18.18} & \textit{0.8224} \\
    Feed Forward & 35.31 & 35.17 & 0.8162 & \textbf{13.25} & 0.6174 & 18.21 & 0.8500 \\
    Attention & 36.23 & 34.49 & 0.7986 & 14.95 & 0.6095 & 19.24 & 0.8644 \\
    Embeddings & 36.13 & 34.30 & 0.7232 & 16.65 & 0.5902 & 19.29 & 0.8192 \\
    Non-Embedding & \textbf{34.57} & \textbf{37.24} & \textbf{0.8760} & \textbf{13.46} & \textbf{0.7991} & \textbf{17.37} & \textbf{0.8845} \\
    \textit{All} & \textit{33.51} & \textit{38.89} & \textit{0.8759} & \textit{5.79} & \textit{0.7999} & \textit{15.75} & \textit{0.8845} \\
    \bottomrule
\end{tabular}
\caption{\textbf{We can improve performance on a target time by swapping out weights with a time vector finetuned on that time.} T5-small start-year finetuned model performance on the end-year split for each task (e.g. finetuning on 2015 for PoliAff and evaluating on 2020). We compare the baseline start-year model (none swapped) to versions with various parameter weights from the target-year model, and the target-year model itself (all swapped).}
\label{tab:swapping}
\end{table*}

\begin{figure*}
    \centering
    \includegraphics[width=\textwidth]{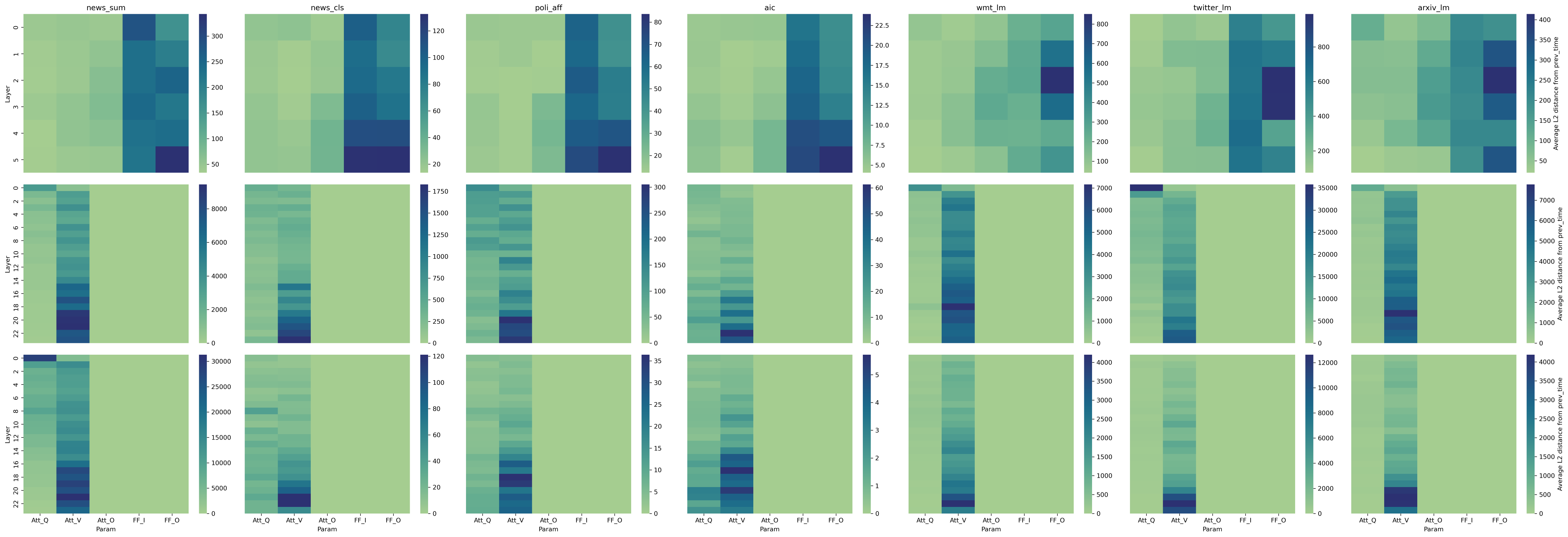}
    \caption{Year-to-year, T5-small feed forward layers change the most across all tasks and domains, and attention changes more in the language modeling setting. For our T5-large and T5-3b models trained with LoRA, the V attention layers change more than the Q layers, with most of the changes (regardless of model size) concentrated in the last layers. Like our param swapping experiment, this suggests that some parameters play a larger role in temporal misalignment than others.}
\end{figure*}

\end{document}